%% file: main.tex
\documentclass[letterpaper,journal]{IEEEtran}
\usepackage{amsmath,amsfonts}
\usepackage{algorithmic}
\usepackage{algorithm}
\usepackage{array}
\usepackage[caption=false,font=normalsize,labelfont=sf,textfont=sf]{subfig}
\usepackage{textcomp}
\usepackage{stfloats}
\usepackage{url}
\usepackage{verbatim}
\usepackage{graphicx}
\usepackage[table]{xcolor}
\usepackage{cite}
\usepackage{pifont}
\usepackage{hyperref}
\usepackage[capitalise]{cleveref}
\usepackage{multirow}
\usepackage{booktabs}

\newcommand{\benchmark}{HeadSwapBench}
\newcommand{\newmethod}{DirectSwap}

\newcommand{\TrainNum}{20,278}
\newcommand{\TestNum}{1,040}
\begin{document}

\title{DirectSwap: Paired, Mask-Free Video Head Swapping with Full-Reference Evaluation}

\author{
Yanan Wang,
Shengcai Liao,~\IEEEmembership{Fellow,~IEEE},
Panwen Hu,
Xin Li,
Fan Yang,
Guangyi Liu,
and Xiaodan Liang%
\IEEEcompsocitemizethanks{
\IEEEcompsocthanksitem Yanan Wang is with Mohamed bin Zayed University of Artificial Intelligence (MBZUAI), UAE.
\IEEEcompsocthanksitem Shengcai Liao is with United Arab Emirates University (UAEU), Al Ain, UAE (e-mail: scliao@uaeu.ac.ae). He is the corresponding author.
\IEEEcompsocthanksitem Panwen Hu, Xin Li, Fan Yang and Guangyi Liu are with Mohamed bin Zayed University of Artificial Intelligence (MBZUAI), Abu Dhabi, UAE.
\IEEEcompsocthanksitem Xiaodan Liang is with Mohamed bin Zayed University of Artificial Intelligence (MBZUAI), Abu Dhabi, UAE, and Sun Yat-sen University (SYSU), Guangzhou, China.
}}


\maketitle
\input{sec/0_abstract}


\input{sec/1_intro}

\input{sec/2_related}

\input{sec/3_dataset}

\input{sec/4_benchmark}

\input{sec/5_method}
\input{sec/6_experiment}
\input{sec/7_conclusion}

\section*{Acknowledgments}
The authors would like to thank all participants in the user study for their time, careful evaluation, and valuable feedback. A large language model was used for language polishing and to assist with some evaluation scripts; see Appendix~I of the supplementary material.

\vfill
\bibliographystyle{IEEEtran}
\bibliography{main}
\clearpage
\appendix
\input{sec/supply}
\end{document}

%% file: sec/0_abstract.tex
\begin{abstract}

Head swapping replaces an entire head while preserving pose, expression, body motion, and scene. Progress is limited by the lack of cross-identity paired videos: real footage cannot provide different identities performing exactly the same motion, leaving the task without paired supervision or frame-aligned ground truth. Existing methods therefore rely on same-identity masked reconstruction, which restricts supervision to predefined editable regions. To address this, we introduce an identity-expression decoupled synthesis pipeline that constructs expression-synchronized cross-identity video pairs from real footage. Expression-bearing facial regions are retained under small clip-consistent geometric perturbations, while the surrounding head is regenerated with a new identity as synthesized swapping input. 
These pairs form a cross-identity benchmark with frame-aligned real-world video as ground-truth targets, enabling full-reference evaluation of identity, expression, pose, reconstruction fidelity, and temporal stability. This yields HeadSwapBench, the first cross-identity paired dataset for video head swapping, supporting both training (\TrainNum{} videos) and benchmarking (\TestNum{} videos). With the cross-identity supervision enabled by this paired dataset, we propose DirectSwap, a mask-free video head swapping training paradigm. Under otherwise identical settings, this formulation outperforms head- and rectangle-masked same-identity reconstruction, particularly for bidirectional head-silhouette changes. At inference, driving-output divergence and emergent reference attention estimate the edit support, allowing unchanged non-head content to be restored from the driving video without external segmentation or additional training. On 1,040 clips from 104 unseen subjects, the resulting model demonstrates that the proposed paired supervision supports effective whole-head swapping. Data, benchmark, evaluation code, and models will be released for research use at \url{https://github.com/Yanan-Wang-cs/DirectSwap}.

\end{abstract}

%% file: sec/1_intro.tex
\section{Introduction}
\label{sec:intro}
\begin{figure*}[t]
  \centering
  \includegraphics[width=\linewidth]{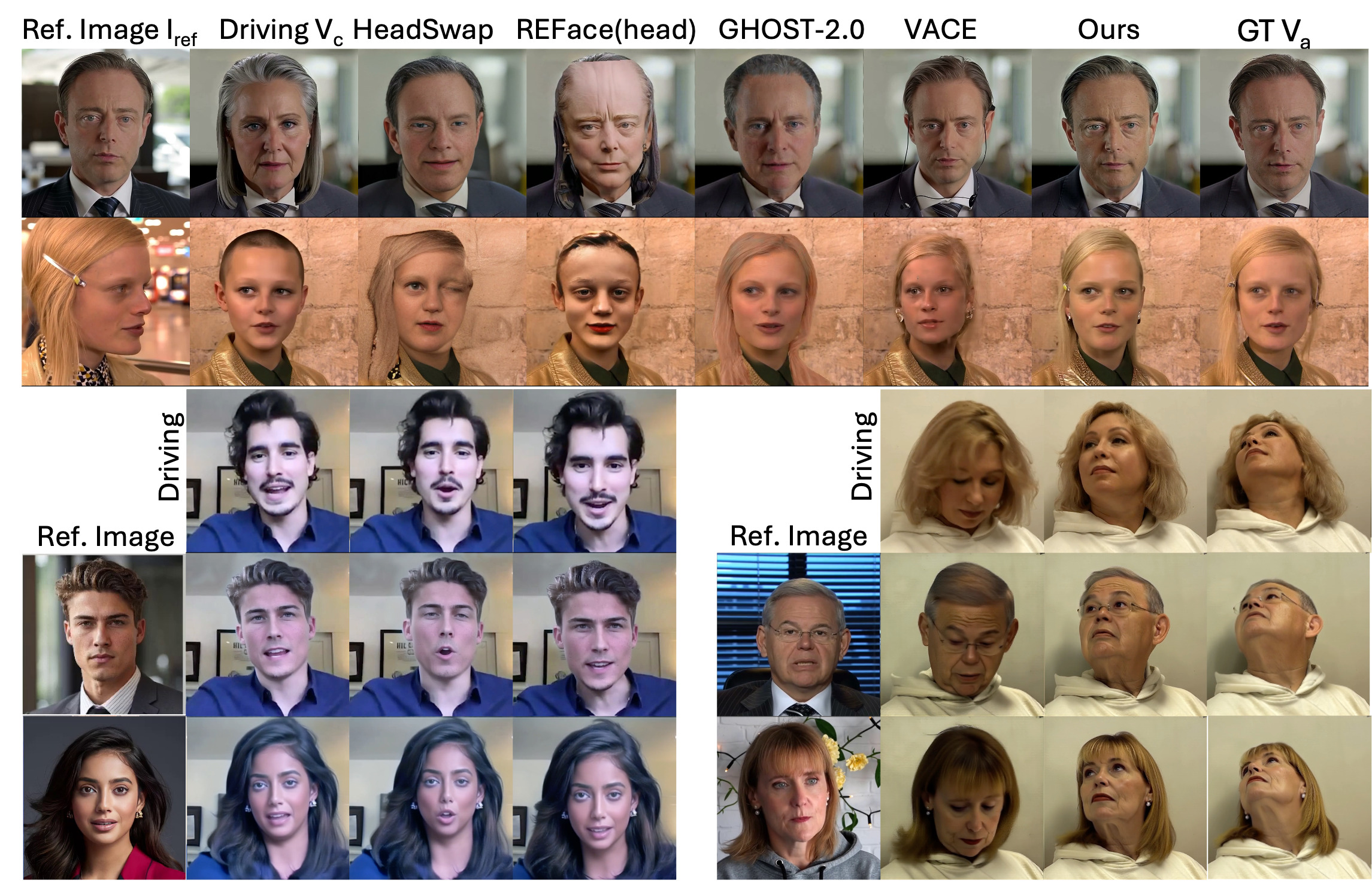}
  \caption{\textbf{Qualitative results for one-shot video head swapping.} \textbf{Top:} Comparison with existing methods on \benchmark{}, where paired ground-truth videos enable direct evaluation. \textbf{Bottom:} Temporally consistent results across frames produced by our video-DiT framework.}
  \label{fig:headswap}
\end{figure*}

Face swapping~\cite{kim2025diffface,chen2020simswap,Gao_2021_CVPR, zhao2023diffswap,baliah2024realistic} transfers facial identity while largely retaining the driving hairstyle, head shape, and outer facial geometry. This restriction places an intrinsic limit on how completely a person's appearance can be transferred (\cref{fig:faceswap}). Head swapping instead replaces the entire head, including the face, hair, ears, jawline, and head geometry, while preserving the driving video's pose, expression, body motion, and scene. The task therefore turns a localized facial edit into a holistic video-generation problem that must jointly model identity, geometry, and temporal facial dynamics. It has applications in cinematic post-production, telepresence, personalized content creation, and privacy-preserving media generation.

Representative head-swapping methods explore head reenactment, diffusion synthesis, mask design, and landmark retargeting~\cite{shu2022few,wang2022hs,groshev2025ghost2,kang2025zero,ji2025controllable}. Despite these advances, they share a limitation: real footage cannot provide different identities performing exactly the same motion in the same scene. Consequently, video head swapping lacks cross-identity paired supervision and frame-aligned ground truth. Two consequences follow.

\begin{figure}[t!]
  \centering
\includegraphics[width=\linewidth]{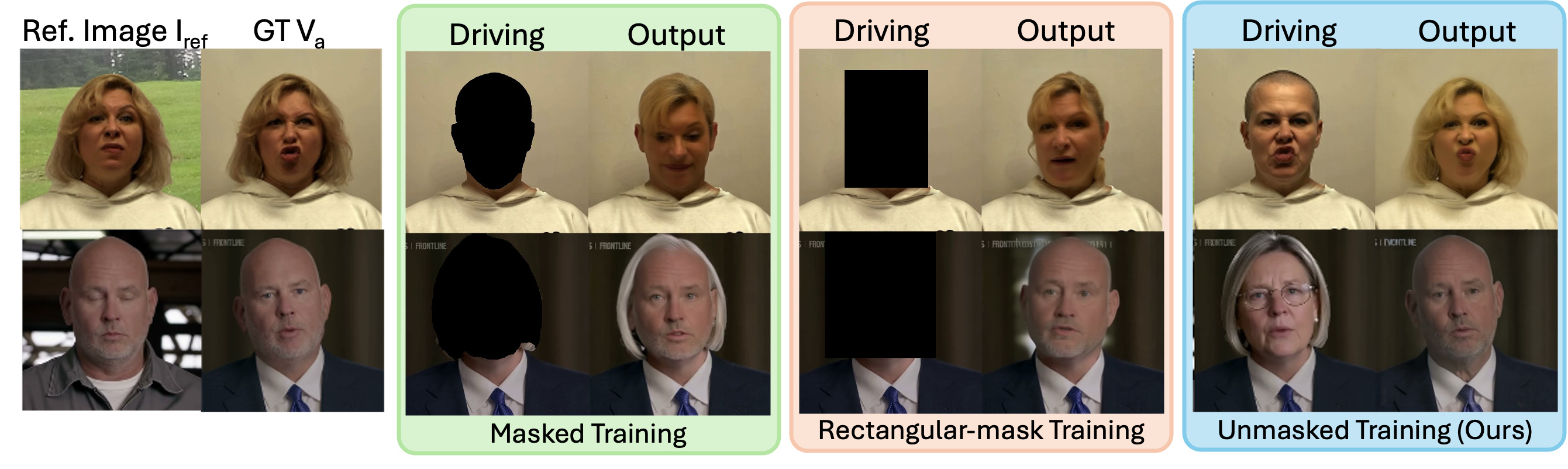}
  \caption{\textbf{Effect of the training input.} 
  Tight head masks constrain geometric changes, whereas rectangular masks remove expression cues and useful background context near head. Our full-frame formulation avoids both constraints.}
  \label{fig:datavalid}
\end{figure}

First, without cross-identity supervision, approaches are typically trained with same-identity reconstruction and formulate head swapping as mask-guided inpainting or compositing. As illustrated in \cref{fig:datavalid}, this creates an inherent spatial trade-off: a tight mask preserves surrounding context but confines generation to the driving-head silhouette, whereas a larger mask provides more freedom at the cost of removing neck, clothing, and background evidence. More fundamentally, same-identity masked reconstruction provides no direct supervision for bidirectional silhouette changes, because the masked input and reconstruction target share the same head geometry. The model is therefore not taught either to generate beyond the driving-head silhouette or to recover content in regions vacated by a smaller target head. Enlarging the mask at inference cannot compensate for this missing cross-identity supervision.

Second, without a frame-aligned target, existing protocols cannot directly evaluate the desired head-swapped video. Identity is measured against a single reference image, but pose mismatch can confound similarity scores: under large pitch mismatch, a same-identity frontal reference can even score below a different-identity reference (\cref{fig:eva}; additional examples in the supplementary material). Background preservation and frame-wise fidelity are difficult to assess without ground-truth video, motivating frame-aligned full-reference evaluation.

\begin{figure}
  \centering
  \includegraphics[width=0.8\linewidth]{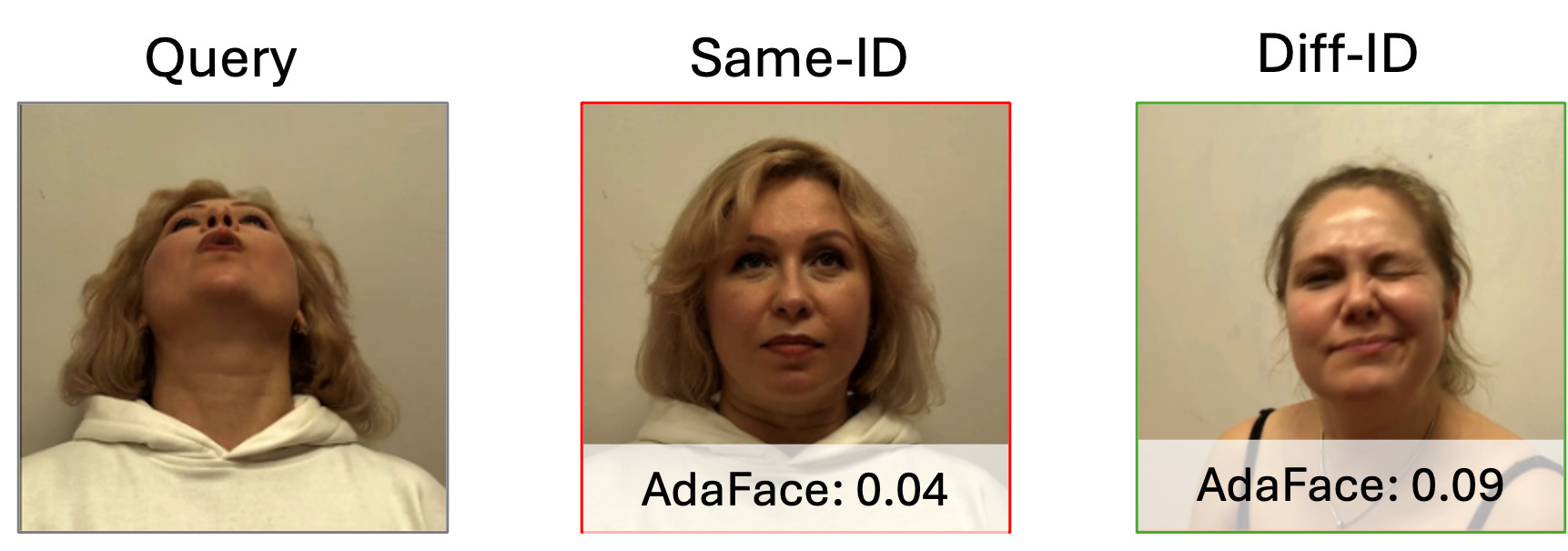}
  \caption{Pose mismatch can confound reference-image identity evaluation.}
  \label{fig:eva}
\end{figure}

To address both limitations, we introduce an identity-expression decoupled pipeline for synthesizing paired cross-identity videos from real footage. Motivated by parametric face models that separate identity and expression factors~\cite{blanz2023morphable,li2017flame}, we preserve localized expression-bearing facial components, including the eyebrows, eyes, nose, and mouth, while applying small geometric perturbations to their layout. A video inpainting model then regenerates the surrounding face, head geometry, texture, and hairstyle with a new identity. The synthesized different-identity video is used as the driving input, while the original real video serves as the frame-aligned target, yielding cross-identity supervision with real-video ground truth. Accepted pairs are filtered for identity separation and facial-dynamics consistency.

The paired data enable full-reference evaluation. Each sample contains a synthetic cross-identity driving video $V_c$, a target-identity reference $I_{ref}$, and a synchronized real target $V_a$, where $V_c$ and $V_a$ share pose, expression, body motion, and scene but differ in whole-head identity. Using these tuples, we construct \benchmark{}, a cross-identity benchmark with frame-aligned video targets. Unlike reference-image evaluation under potential pose mismatch, \benchmark{} compares each output with the corresponding frame in $V_a$.

The paired data enable a direct formulation of head swap training without predefined editable regions. We instantiate this mask-free formulation with DirectSwap, which learns directly from $(V_c,I_{ref},V_a)$ without head or region masks. Unlike mask-guided inpainting, DirectSwap receives the complete driving video $V_c$, whose visible head provides pose and expression cues, while $I_{ref}$ specifies the target identity, head shape, and hairstyle. The synchronized target video $V_a$ directly supervises both directions of silhouette change: recovery of background, neck, or clothing when the target head is smaller, and generation beyond the driving-head silhouette when it is larger. DirectSwap can therefore learn bidirectional silhouette changes without being confined to a predefined driving-head region.

Mask-free full-frame generation removes spatial constraints on the head, but may also alter unchanged non-head content, including background, neck, and clothing. Our key idea is to identify the regions affected by the swap and preserve generated content only within those regions, while restoring the remaining content from the driving video. Specifically, we derive a driving-output divergence map to capture changed regions, including areas vacated by the driving head, and an emergent reference-attention map to localize the generated target-head region. Their union defines the edit support used to restore unchanged non-head content from the driving video. This training-free procedure requires no external segmentation, additional training, or trainable parameters.

In summary, our contributions are threefold:
\begin{itemize}
    \item We introduce an \textbf{identity-expression decoupled data synthesis pipeline} that constructs cross-identity paired videos by preserving expression-bearing facial regions while regenerating the surrounding head with a different identity, followed by automatic filtering for identity separation and facial-dynamics consistency.
        
    \item We construct \textbf{\benchmark{}, a paired and cross-identity video head-swapping benchmark} with frame-aligned real-video targets, enabling direct full-reference evaluation of the desired head-swapped video.
        
    \item We introduce \textbf{DirectSwap} to exploit the proposed paired supervision through mask-free cross-identity training for video head swapping. Under otherwise identical settings, this formulation outperforms head- and rectangle-masked reconstruction, while driving-output divergence and reference attention enable training-free preservation of unchanged non-head content at inference.
    
\end{itemize}

%% file: sec/2_related.tex
\section{Related Work}
\label{sec:related}

\noindent\textbf{Head Swapping Methods.}
Head swapping remains underexplored compared with face swapping. Most conventional face-swapping methods~\cite{kim2025diffface,zhao2023diffswap} focus on the facial region while largely preserving surrounding hair and head context, thereby limiting holistic transfer of hairstyle and head morphology. HeSer~\cite{shu2022few} and GHOST-2.0~\cite{groshev2025ghost2} formulate whole-head transfer through head reenactment followed by scene/background blending, while HS-Diffusion~\cite{wang2022hs} combines semantic-layout mixing with latent diffusion, semantic calibration, and neck alignment. HID~\cite{kang2025zero} introduces the context-aware IOMask, and SwapAnyHead~\cite{ji2025controllable} uses shape-agnostic masks and landmark retargeting for controllable video head swapping. Despite architectural differences, many of these methods rely on explicit spatial partitioning to determine where the transferred head should be synthesized or composited, whereas our method learns whole-head transfer from paired cross-identity supervision without requiring an explicit head mask.

\noindent\textbf{Synthetic Supervision for Identity Swapping.}
A separate line of work addresses the absence of cross-identity ground truth by constructing synthetic paired supervision. ReliableSwap~\cite{yang2023reliableswap} constructs cycle triplets via face reenactment and blending, using synthesized swapped faces as training inputs and real images as reliable pixel-level supervision. DreamID~\cite{ye2025dreamid} constructs Triplet ID Groups from two same-identity images and a pseudo target obtained by swapping a different identity onto one of them, with the original pre-swap image serving as pixel-level ground truth. PixSwap~\cite{kim2025pixswap} similarly constructs synthetic paired data with pseudo ground truth to provide direct pixel-level supervision for identity transfer and target-attribute preservation. LivingSwap~\cite{luo2026preserving} extends synthetic paired supervision to video by applying per-frame face swapping and reversing the resulting data pairs, using the edited clip as input and the original video as ground truth. Concurrently, DreamID-V~\cite{guo2026dreamidv} combines image face swapping with an identity-anchored, pose-conditioned video synthesizer to construct bidirectional identity quadruplets, and introduces IDBench-V for video-level evaluation. These approaches construct paired supervision through proxy swapping or conditioned synthesis. Our framework instead imposes identity-expression separation directly in the synthesized video pair: expression-bearing facial appearance is preserved from the real video under small geometric perturbations, while the surrounding head geometry, texture, and hairstyle are regenerated with a new identity. The resulting pairs support whole-head transfer with explicit cross-identity supervision and provide frame-aligned real-video targets for evaluation.

\noindent\textbf{Background Preservation in Diffusion Editing.}
Existing methods differ mainly in how editable regions are obtained. Mask-guided approaches rely on externally specified regions and preserve the complement through source blending or background key-value reuse~\cite{kvedit}. DiffEdit~\cite{diffedit} instead derives masks from differences between denoising predictions, while InstDiffEdit~\cite{instdiffedit} constructs masks from text-image cross-attention. More recent instruction-based localization methods further exploit DiT attention and latent features to identify edit regions~\cite{he2026rethinking}, while mechanistic analyses show that instance identity can be transmitted directly from the reference image to generated content through image-to-image attention~\cite{ge2026binding}. Our method turns this reference-to-output attention into a spatial read-out: without an externally specified or text-grounded edit mask, we localize the generated reference-conditioned subject in video and combine this signal with driving-output divergence to capture both the generated head region associated with the reference identity and regions vacated by the original driving head, restoring only the unaffected complement through training-free compositing.

%% file: sec/3_dataset.tex
\section{Identity-Expression Decoupled Synthesis}
\label{sec:dataset1}

\begin{figure*}[t!]
  \centering
  \includegraphics[width=\linewidth]{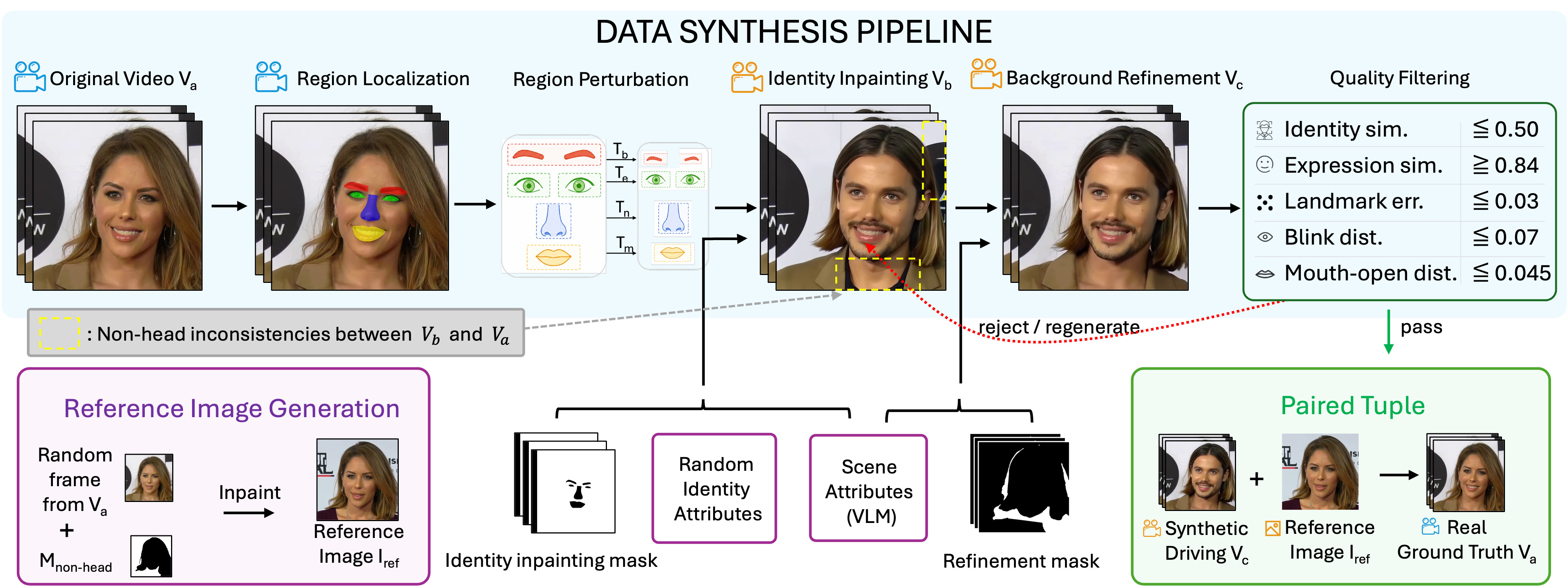}
  \caption{\textbf{Identity-expression decoupled data synthesis pipeline.} Given a real video $V_a$, we perturb expression-bearing facial regions and regenerate the surrounding head with a new identity to obtain $V_b$. Dashed boxes mark non-head artifacts introduced by identity inpainting. We then composite the synthesized head with the original scene and refine residual gaps to produce $V_c$. A reference image $I_{ref}$ is generated from a random frame. After filtering, accepted samples form tuples $(V_c,I_{ref},V_a)$, with $V_a$ serving as frame-aligned ground truth.}
  \label{fig:datasetoverview}
\end{figure*}

Video head swapping lacks cross-identity supervision because real footage cannot provide two identities with identical motion and scene. We introduce an identity-expression decoupled synthesis pipeline that converts videos into tuples $(V_c,I_{ref},V_a)$, consisting of a cross-identity driving video, a target-identity reference, and a synchronized real target.

Our key idea is to impose identity-expression separation structurally during data synthesis. We preserve localized facial regions carrying expression-dependent appearance, perturb their relative geometry, and regenerate the surrounding face, head geometry, texture, and hairstyle with a new identity. The resulting pairs support both the full-reference evaluation in \cref{sec:benchmark} and the mask-free training in \cref{sec:method}.

\subsection{Data Synthesis Pipeline}
\label{sec:data-pipeline}
As shown in \cref{fig:datasetoverview}, the pipeline has a video-synthesis branch and a reference-image branch. The former performs region localization, expression-preserving perturbation, identity inpainting, background refinement, and quality filtering, while the latter generates the target-identity reference $I_{ref}$.

\noindent\textbf{Original Video $V_a$.}
We begin with single-person real videos collected from HDTF~\cite{zhang2021flow}, VFHQ~\cite{xie2022vfhq}, and FEED~\cite{drobyshev2024emoportraits}, covering talking-head speech, high-quality near-frontal interviews, and large-pose expressive motions, respectively. Each video is denoted by $V_a$ and retained as the ground-truth target.

\noindent\textbf{Region Localization.}
We localize expression-bearing regions in each frame of $V_a$ using FaRL~\cite{zheng2022farl}, including the eyebrows, eyes, nose, and mouth. These regions are retained to preserve expression-related appearance during
identity synthesis.

\noindent\textbf{Region Perturbation.}
The six localized regions are grouped into four perturbation units: the two eyebrows, the two eyes, the nose, and the mouth. Each unit undergoes a small affine transformation $T_b$, $T_e$, $T_n$, or $T_m$ with random scaling and translation, sampled once per video and fixed across frames for temporal consistency. For the eyebrow and eye pairs, both sides share the same scale and vertical translation, with horizontal translations mirrored about the facial midline; the nose and mouth are perturbed independently. This preserves local expression-related appearance while introducing small, temporally consistent changes to spatial configuration.

\noindent\textbf{Identity Inpainting $V_b$.}
We construct an identity-inpainting mask from the bounding rectangle of the original head by first expanding the box outward by 10\% on all sides and then extending its lower boundary downward by an additional 50\% of the head height, while excluding the perturbed facial units. This expanded region provides sufficient freedom to modify facial shape, head geometry, and hairstyle, whereas the retained facial regions preserve expression-related appearance.

We perform video inpainting with Wan2.1-T2V-14B~\cite{wan2025} using a ComfyUI workflow~\cite{comfyui}. The model is conditioned on randomly sampled identity attributes and scene attributes extracted by Qwen2.5-VL-72B-Instruct~\cite{qwen2.5-VL}. The identity prompt specifies an age group (\emph{teenage}, \emph{young}, \emph{middle-aged}, or \emph{elderly}), gender (\emph{man} or \emph{woman}), body build (\emph{heavyset}, \emph{slim}, or \emph{medium-build}), hair length (\emph{long}, \emph{medium-length}, or \emph{short}), and hairstyle (\emph{straight} or \emph{curly}). The scene attributes describe the original background and lighting. The resulting intermediate video $V_b$ depicts a new head identity while retaining expression cues from the preserved facial regions. The expanded inpainting region may introduce unintended changes around the
hair, neck, shoulders, or background, as shown by the dashed boxes in
\cref{fig:datasetoverview}.

\noindent\textbf{Background Refinement $V_c$.}
To remove the non-head inconsistencies in $V_b$, we composite the synthesized head with the original body and scene from $V_a$. To handle gaps or boundary artifacts caused by silhouette differences, we construct a refinement mask around the hairline, neck, and shoulders and apply a second video-inpainting step, using the same video-inpainting workflow as above. This produces the final synthetic driving video $V_c$, which retains the new head identity while recovering the original non-head content.

\noindent\textbf{Quality Filtering.}
Each candidate $V_c$ is compared with its synchronized real target $V_a$ using five automatic criteria, with thresholds chosen by visual inspection. A candidate is rejected and regenerated when its identity similarity exceeds $0.50$, its expression similarity falls below $0.84$, its normalized landmark error exceeds $0.03$, its blink distance exceeds $0.07$, or its mouth-opening distance exceeds $0.045$. Identity similarity is computed using AdaFace~\cite{kim2022adaface}, while expression, landmarks, blinking, and mouth opening are measured using MediaPipe~\cite{lugaresi2019mediapipe}. Approximately 78\% of candidates pass all criteria, removing samples with excessive identity similarity or inconsistent facial dynamics. Appendix~A of the supplementary material compares keypoint-conditioned regeneration, for which only 0.4\% of candidates pass the same filter.

\noindent\textbf{Reference Image Generation.}
In parallel with video synthesis, we generate a reference image $I_{ref}$ for conditioning the head-swapping model. A frame is randomly sampled from $V_a$, and a head mask $M_{\mathrm{head}}$ is obtained by fusing the segmentation results of SCHP~\cite{li2020schp} and FaRL~\cite{zheng2022farl}. We define the complementary non-head mask as $M_{\mathrm{non\text{-}head}} = \mathbf{1}-M_{\mathrm{head}}$, retain the head region, and inpaint the non-head region with the same video-inpainting workflow to produce $I_{ref}$. The resulting reference image preserves the target head appearance while removing potentially distracting non-head context, and is used together with the synthetic driving video $V_c$.

\noindent\textbf{Final Paired Tuple.}
Each accepted sample forms a tuple $(V_c,I_{ref},V_a)$, providing direct full-frame supervision for mask-free training and frame-aligned real targets for full-reference evaluation.

\subsection{Comparison of Paired-Synthesis Strategies}
\label{sec:data-comparison}

Existing synthetic paired-supervision methods construct cross-identity pairs through different synthesis strategies, most commonly using pretrained swapping or reenactment models~\cite{yang2023reliableswap,ye2025dreamid, kim2025pixswap,luo2026preserving,guo2026dreamidv}. \cref{tab:related_comparison} compares these methods in terms of data type, pair-construction strategy, and public availability. Unlike prior approaches, our pipeline does not transfer a prescribed reference identity onto the driving content. Instead, it perturbs expression-bearing facial regions and regenerates the surrounding head with a new identity, yielding cross-identity video pairs whose supervision covers the entire head rather than only facial identity.

\begin{table}[t]
\centering
\small
\setlength{\tabcolsep}{2.5pt}
\caption{Comparison of synthetic paired supervision for identity swapping. ``I''/``V'' denote image/video, and ``Construction'' denotes generation method. DreamID-V only releases the benchmark, while ours releases both training and benchmark.}
\label{tab:related_comparison}
\begin{tabular}{lccc}
\hline
Method & Type & Construction & Release \\
\hline
ReliableSwap~\cite{yang2023reliableswap}& Face-I& Reenactment + Blend& \ding{55}\\

DreamID~\cite{ye2025dreamid}& Face-I& Face Swap& \ding{55}\\

PixSwap~\cite{kim2025pixswap}& Face-I& Face Swap& \ding{55}\\

LivingSwap~\cite{luo2026preserving}& Face-V& Frame-wise Swap& \ding{55}\\

DreamID-V~\cite{guo2026dreamidv}& Face-V& Face Swap + Video Synth.& \checkmark\\

\textbf{Ours}
& \textbf{Head-V}& \textbf{Region Perturb. + Inpaint}& \checkmark\\
\hline
\end{tabular}
\end{table}

%% file: sec/4_benchmark.tex
\section{Full-Reference Benchmark}
\label{sec:benchmark}
Using the paired data constructed in \cref{sec:dataset1}, we build \benchmark{}, a full-reference benchmark for video head swapping. For each tuple $(V_c, I_{ref}, V_a)$, a method takes $V_c$ and $I_{ref}$ as input, and the synchronized real video $V_a$ serves as the frame-aligned evaluation target.

\subsection{Benchmark Construction and Cross-Identity Splits}
\label{sec:benchmark-split}

After synthesis and quality filtering, we partition the accepted pairs into cross-identity training, validation, and test sets. We first group videos according to the identity annotations provided by the original datasets, treating videos with the same identity ID as belonging to the same subject. Since different IDs may still correspond to the same person, we further compare the AdaFace~\cite{kim2022adaface} embeddings of the reference image associated with each identity across different IDs to identify duplicate subjects. The resulting identity groups are assigned exclusively to one of the three splits, ensuring no subject overlap among training, validation, and test sets.

This procedure yields 20,278 training clips, 40 validation clips from 16 subjects, and 1,040 test clips from 104 subjects. All clips are $512{\times}512$ at 16\,fps and contain up to 121 frames ($\approx$7.6\,s; median 121 and minimum 53 frames on the test split). The validation split is reserved exclusively for checkpoint selection and calibration of the inference-time read-out in \cref{sec:bgpreserve}; no threshold or design choice is tuned on the test split. It contains only HDTF and VFHQ clips, so calibration is performed without access to the large-expression and large-pose FEED footage included in the test set. For benchmark reliability, we manually inspect all 1,040 test samples and verify that the reference image $I_{ref}$ and the synchronized ground-truth video $V_a$ depict the same target identity.

The cross-identity protocol ensures that evaluation reflects generalization to unseen subjects rather than familiarity with identities observed during training. Because the number of clips varies across subjects, we further adopt per-subject aggregation for evaluation, as described in \cref{sec:metrics}.

\subsection{Evaluation Protocol}
\label{sec:metrics}

Because different subjects contribute different numbers of clips, directly averaging over all clips would assign larger weights to identities with more observations. We therefore adopt per-subject aggregation. Unless otherwise stated, frame-level measurements are first averaged within each clip, the resulting clip-level scores are then averaged over all clips of the same subject, and the final result is obtained by averaging across test subjects. This gives each unseen identity equal weight in the reported performance.

\noindent\textbf{Why Frame-Synchronized Comparison.}
Face- and head-embedding similarities can become unreliable under large pitch mismatch. As shown in \cref{fig:eva} and Appendix~D, a same-identity frontal reference can receive similarity scores comparable to those of different identities, whereas pose-matched same-identity comparison restores clear separation. Thus, evaluation against a single reference image can confound identity fidelity with viewpoint mismatch. Our full-reference benchmark instead compares each generated frame with the synchronized target frame in $V_a$, substantially reducing this effect and providing a more reliable measure.

\noindent\textbf{Identity Similarity ($\mathit{Face}$).}
Facial identity is measured by the cosine similarity between AdaFace~\cite{kim2022adaface} embeddings extracted from each generated frame and its corresponding frame in $V_a$. 

\noindent\textbf{Head Similarity ($\mathit{Head}$).}
Face-recognition embeddings focus on aligned facial crops and underrepresent hairstyle and head geometry. We therefore report whole-head similarity using the structure-aware embedding of~\cite{wang2026head}, computed between generated frames and synchronized target frames in $V_a$ over the full head to capture appearance beyond the cropped face.

\noindent\textbf{Pose and Expression Consistency.}
We use MediaPipe Face Landmarker~\cite{lugaresi2019mediapipe} to extract expression blendshape coefficients, head pose, and facial landmarks from the generated video and $V_a$. Expression similarity ($\mathit{Exp}$) is computed as the cosine similarity between blendshape coefficients. Pose error ($\mathit{Pose}$) is the mean squared error of mean-centered pitch, yaw, and roll angles, reducing static bias from differences in head geometry. Keypoint error ($\mathit{Key}$) measures the normalized distance between corresponding landmarks. Blink distance ($\mathit{Eye}$) and mouth-open distance ($\mathit{Mouth}$) are L1 distances computed from the corresponding eye-aperture and mouth-opening signals.

\noindent\textbf{Reconstruction and Perceptual Fidelity.}
Paired real targets enable direct full-reference measurements that are not available under conventional unpaired evaluation. We report SSIM~\cite{wang2004image}, PSNR~\cite{huynh2008scope}, and LPIPS~\cite{zhang2018unreasonable} between each generated frame and its corresponding frame in $V_a$. LPIPS is computed using AlexNet~\cite{krizhevsky2012imagenet}. We additionally report FID~\cite{heusel2017gans} between generated and real test-frame distributions as a distribution-level measure of visual realism.

\noindent\textbf{Temporal Stability.}
We report tLPIPS~\cite{siarohin2019first} to measure variation across adjacent frames. Lower values indicate greater temporal consistency and reduced frame-to-frame flicker.

\subsection{Dataset Statistics and Coverage}
\label{sec:benchmark-analysis}

\noindent\textbf{Cross-Identity Pair Validity.}
A valid full-reference pair should differ in head identity while remaining aligned in motion and scene attributes preserved by head swapping. As shown in \cref{tab:quality_stats}, low identity similarity between $V_c$ and $V_a$ confirms identity separation, while high expression similarity and low pose, landmark, blink, and mouth-opening errors indicate aligned face dynamics. These results verify cross-identity paired evaluation rather than same-identity reconstruction.

\begin{table}[t]
\centering
\small
\setlength{\tabcolsep}{2.5pt}
\caption{Validity of cross-identity pairs in \benchmark{}. Face, Head, and Exp are similarities; Key, Eye, Mouth, and Pose are errors.}
\label{tab:quality_stats}
\begin{tabular}{lcccccccc}
\hline
&Clips & $\mathit{Face}$ & $\mathit{Head}$ & $\mathit{Exp}$ & $\mathit{Key}$ & $\mathit{Eye}$ & $\mathit{Mouth}$ & $\mathit{Pose (deg^2)}$ \\
\hline
Train & 20278 & 0.22 & 0.43 & 0.87 & 0.03 & 0.05 & 0.03 & 0.95 \\
Val   & 40    & 0.22 & 0.44 & 0.87 & 0.03 & 0.05 & 0.03 & 1.10 \\
Test  & 1040  & 0.22 & 0.45 & 0.87 & 0.03 & 0.05 & 0.03 & 1.12 \\
\hline
\end{tabular}
\end{table}

\noindent\textbf{Expression and Motion Coverage.}
Beyond pairwise synchronization, a useful benchmark should cover diverse facial dynamics and challenging head motion. \cref{tab:coverage} summarizes expression diversity, speech activity, and large head rotations in the real target videos $V_a$. The test split covers all eight expression categories: mouth open, surprise, frown, smile, brow raise, pucker, squint, and eyes closed. Only 6.0\% of clips trigger none of these categories and are classified as neutral, while 121 distinct co-occurrence patterns are observed, demonstrating substantial diversity in compound facial expressions. We report these coverage statistics for the training and test splits; the small validation split is used only for model selection and inference-time calibration.

Using MediaPipe Face Landmarker, we estimate 52 per-frame blendshape scores in $[0,1]$. Excluding the neutral coefficient, we define a clip as exhibiting a \emph{strong} (\emph{extreme}) expression if any non-neutral blendshape exceeds $0.8$ ($0.9$) in at least one frame; 87.4\% and 51.2\% of test clips satisfy these criteria, respectively. We define a \emph{large head rotation} as any clip containing a frame with $|\mathrm{yaw}|>30^\circ$ or $|\mathrm{pitch}|>25^\circ$, which occurs in 17.2\% of the test split. Speech statistics, head-pose distributions, and expression-category definitions are provided in Appendix~A of the supplementary material.

\begin{table*}[t]
\centering
\small
\setlength{\tabcolsep}{3.2pt}
\caption{Expression, speech, and pose coverage of \benchmark{}. The Expression Categories report the percentage of clips in which each expression is detected, or none of them (Neutral), while Strong and Extreme summarize expression intensity. Speech and Large-Pose report percentage of clips having speech activity and large head rotations, respectively.}
\label{tab:coverage}
\begin{tabular}{l|ccccccccc|cc|c|c}
\hline
& \multicolumn{9}{c|}{Expression Categories}
& \multicolumn{2}{c|}{Expression Intensity}
& \multirow{2}{*}{Speech}
& \multirow{2}{*}{Large Pose} \\
& Mouth-Open
& Surprise
& Frown
& Smile
& Brow-Raise
& Pucker
& Squint
& Eyes-Closed
& Neutral
& Strong
& Extreme
& 
&  \\
\hline
Train
& 26.5\% & 14.2\% & 35.6\% & 46.9\% & 28.2\% & 40.8\% & 61.8\% & 16.6\%
& 4.9\%  & 82.8\% & 47.9\%& 61.6\% & 17.4\% \\
Test
& 29.7\% & 22.2\% & 33.0\% & 43.4\% & 37.6\% & 42.5\% & 64.7\% & 20.8\%
& 6.0\%  & 87.4\% & 51.2\%& 60.6\% & 17.2\% \\
\hline
\end{tabular}
\end{table*}

\noindent\textbf{Silhouette-Change Coverage.}
Whole-head swapping is challenging when the target head extends beyond or contracts within the driving-head silhouette. We quantify silhouette change using the normalized hair-length difference $d_{\mathrm{hair}}$ between driving and target heads, where hair length is estimated from the head-segmentation mask. We regard $|d_{\mathrm{hair}}|>0.2$ as a substantial transition. In the test split, 7.3\% of clips (76 clips, 14 subjects) require short-to-long transitions, demanding generation beyond the driving-head silhouette, whereas 8.6\% (89 clips, 13 subjects) require long-to-short transitions, which demand recovery of background, neck, or clothing in regions vacated by the driving-head footprint. The corresponding training proportions are 5.9\% and 4.8\%. Overall, approximately 16\% of the test split exercises bidirectional silhouette changes that are difficult to supervise with predefined editable masks; per-method results on these subsets are reported in \cref{tab:silhouette}.

%% file: sec/5_method.tex
\section{\newmethod{} Framework}
\label{sec:method}

\begin{figure*}[pt]
  \centering
  \includegraphics[width=1.0\linewidth]{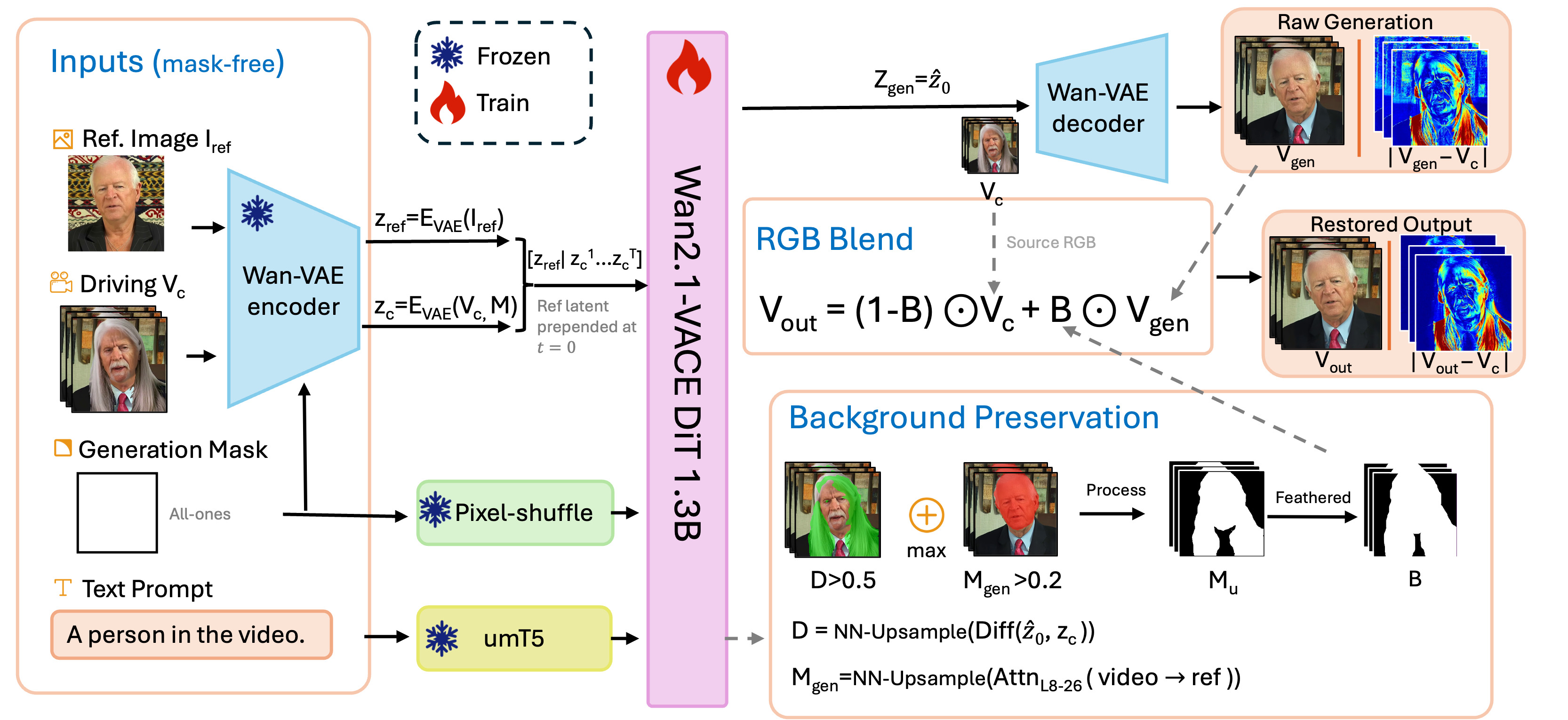}
  \caption{DirectSwap pipeline with mask-free generation and training-free background preservation via $D$ and $A$. The rightmost panels show the absolute differences $|V_{\mathrm{gen}}-V_c|$ and $|V_{\mathrm{out}}-V_c|$ before and after background restoration. NN-Upsample denotes nearest-neighbor upsampling.}
  \label{fig:directswap}
\end{figure*}

Given the paired tuples $(V_c, I_{ref}, V_a)$ introduced in \cref{sec:dataset1}, we develop \textbf{DirectSwap}, a video head-swapping framework that learns from full frames without a predefined mask. As illustrated in \cref{fig:directswap}, $V_c$ provides synchronized driving motion, $I_{ref}$ specifies the target head identity, and $V_a$ serves as the frame-aligned training target. We instantiate DirectSwap with Wan2.1-VACE-1.3B~\cite{wan2025}, a video Diffusion Transformer, while retaining its pretrained Wan-VAE and text encoder. Our changes lie in the training formulation and inference-time background-preservation mechanism: paired supervision enables mask-free full-frame generation, while DiT signals are read out at inference to identify the swap support and restore untouched driving content.

\subsection{Mask-Free Paired Training}
\label{sec:maskfree}

Conventional head-swapping methods typically use same-identity reconstruction. Since the input and target depict the same person, the driving head must be masked to prevent trivial reconstruction, and the mask also defines the editable region. This ties supervision to the driving-head boundary and limits changes that extend beyond or vacate part of it.

DirectSwap instead learns from cross-identity paired videos, where the complete driving head in $V_c$ and the synchronized target head in $V_a$ belong to different identities. The driving head therefore does not need to be removed: the identity mismatch itself provides the supervision for replacing the visible head while preserving the shared pose, expression, body motion, and scene. We feed the complete driving video $V_c$ together with the target-identity reference image $I_{ref}$, and set the VACE generation mask to all ones for every frame. The mask thus carries no spatial information about the head location, shape, or editable boundary.

This full-frame paired supervision directly teaches both directions of head-silhouette change. When the target head is larger than the driving head, the model is supervised to generate beyond the driving-head silhouette. When the target head is smaller, the synchronized target $V_a$ provides the correct background, neck, or clothing content in regions vacated by the driving head. DirectSwap therefore learns where the head boundary should expand or contract from the paired target itself, rather than from a predefined editable mask.

\subsection{Training-Free Background Preservation}
\label{sec:bgpreserve}

Mask-free full-frame generation removes the spatial restriction imposed by predefined editable regions, but may also modify content that should remain unchanged. We therefore introduce an inference-only preservation mechanism that restores unchanged non-head content from the driving video.

\begin{figure}
  \centering
  \includegraphics[width=1.0\linewidth]{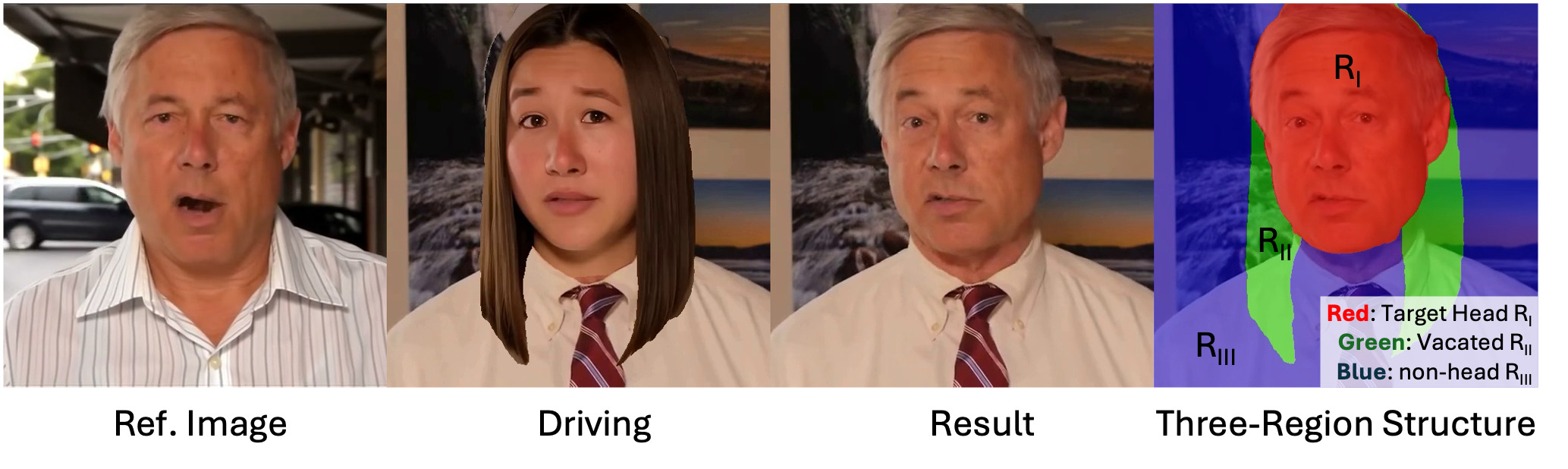}
  \caption{\textbf{Three-region structure of head swapping.}
The scene comprises the generated target head $R_{\mathrm{I}}$, vacated driving-head region $R_{\mathrm{II}}$, and unchanged non-head region $R_{\mathrm{III}}$. Only $R_{\mathrm{III}}$ is restored from the driving video.}
  \label{fig:three}
\end{figure}
\noindent\textbf{Three-Region Structure.}
As illustrated in \cref{fig:three}, a head swap naturally induces three spatial regions: the generated target head $R_{\mathrm{I}}$, the footprint vacated by the driving head $R_{\mathrm{II}}$, and the unchanged non-head region $R_{\mathrm{III}}$. Regions $R_{\mathrm{I}}$ and $R_{\mathrm{II}}$ must retain the generated result, whereas only $R_{\mathrm{III}}$ can be safely restored from $V_c$. In particular, copying $R_{\mathrm{II}}$ from the driving video would reintroduce removed driving-head content. We therefore need to identify the combined edit support $R_{\mathrm{I}}\cup R_{\mathrm{II}}$, outside which content can be restored from the driving video.

\noindent\textbf{Driving-Output Divergence.}
A natural cue for estimating the edit support is the discrepancy between the generated output and the driving input. We therefore compare the predicted clean latent $\hat z_0$ with the driving latent $z_c$ independently within each frame. At each latent location within a frame, we compute the channel-averaged $\ell_1$ difference
\begin{equation}
E=\frac{1}{C}\sum_{n=1}^{C}\left|\hat z_{0,n}-z_{c,n}\right|,
\end{equation}
and spatially smooth $d$ to obtain $\tilde E$. Let $m=\operatorname{median}\tilde E$ denote the spatial median of each frame. We then define the frame-relative divergence response as
\begin{equation}
D=1-\operatorname{sigmoid}\!\left(
\frac{1.2\,m-\tilde E}{0.4\,m}
\right).
\end{equation}

However, divergence alone introduces a threshold trade-off. With a low threshold, $D$ yields an overly broad edit support and little content is restored; with a higher threshold, more driving content is recovered, but low-divergence facial details may also be copied back from $V_c$, reducing target-identity fidelity. This behavior is also observed in \cref{fig:bg_ablation_vis,tab:bg_ablation} and supplementary material. We therefore introduce a complementary cue that explicitly identifies the reference-conditioned target head.

\noindent\textbf{Emergent Reference-Attention Matte.}
Because the reference image is prepended as a latent frame, video tokens can directly attend to reference-image tokens through the DiT self-attention. We observe that this video-to-reference attention forms a spatial response aligned with the generated target head, despite the absence of any head-mask supervision.

Let $\mathcal{R}$ denote the reference tokens. For video token $i$ at layer $\ell$, we compute its reference attention as $a_i^{(\ell)}=\sum_{j\in\mathcal{R}}A_{ij}^{(\ell)}.$ We use layers $\mathcal{L}=\{8,\ldots,26\}$, selected on the validation split for strongest spatial discrimination. For each layer and latent frame, we select the top $30\%$ of tokens by $a_i^{(\ell)}$ and average the resulting binary indicators across $\mathcal{L}$ to obtain $A$. This rank-based aggregation provides clearer spatial separation than averaging raw attention; details are provided in Appendix~C of the supplementary material.

\noindent\textbf{Edit-Support Union.}
The two signals play complementary roles: $D$ captures changed regions, including the vacated driving-head footprint $R_{\mathrm{II}}$, while $A$ protects the reference-conditioned target head $R_{\mathrm{I}}$. We combine them as
\begin{equation}
U=\left(D>0.5\right)\lor\left(A>0.2\right).
\end{equation}
We denoise $U$ by removing isolated and spurious boundary responses, then apply morphological closing and hole filling:
\begin{equation}
M_u =\operatorname{Fill}\!\left(\operatorname{Close}\!\left[\operatorname{Denoise}(U)\right]\right).
\end{equation}
The $M_u$ approximates $R_{\mathrm{I}}\cup R_{\mathrm{II}}$, while its complement corresponds to the restorable unchanged region $R_{\mathrm{III}}$. The threshold-selection procedure, denoising rules, and sensitivity analyses are provided in Appendix~C of the supplementary material.

\noindent\textbf{RGB Background Restoration.}
$M_u$ is converted into a soft matte $B$ by outward Gaussian feathering with $\sigma=8$ pixels as 
\begin{equation}
B=\max\!\left(M_u,\,\operatorname{Feather}\!\left(\operatorname{Dilate}(M_u,2\sigma),\sigma\right)\right).
\end{equation}
This construction keeps $B=1$ throughout the estimated edit support. We then composite the generated video $V_{\mathrm{gen}}$ with the driving video $V_c$ directly in RGB space as 
\begin{equation}
V_{\mathrm{out}}=B\odot V_{\mathrm{gen}}+(1-B)\odot V_c .
\end{equation}
Thus, generated content is retained over the target head and vacated driving-head footprint, while unchanged regions are restored from $V_c$. Feathering details are provided in Appendix~C of the supplementary material.

The preservation mechanism operates only at inference. Both $D$ and $A$ reuse intermediate signals from inference of the trained DirectSwap model; no external segmentation network, additional learned parameters, or further optimization is required.

%% file: sec/6_experiment.tex
\section{Experiments}
\label{sec:experiments}

We evaluate \newmethod{} on the full-reference benchmark introduced in \cref{sec:benchmark}. The experiments answer four questions: (1) whether the paired supervision can support a competitive head-swapping system under frame-aligned full-reference evaluation (\cref{sec:main-comparison}); (2) whether human raters show trends consistent with the benchmark, and whether these trends extend to footage without ground truth (\cref{sec:user-study}); (3) whether full-frame cross-identity training uses the paired supervision more effectively than head- or rectangle-masked reconstruction (\cref{sec:mask-ablation}); and (4) whether the training-free preservation module improves fidelity of unchanged non-head regions without compromising the head (\cref{sec:bg-ablation}). 

\subsection{Experimental Setup}
\label{sec:exp-setup}

Unless otherwise stated, all methods are evaluated on the 1{,}040-clip test split using the metrics and per-subject aggregation protocol defined in \cref{sec:metrics}. Each method receives the same driving video $V_c$ and target-identity reference image $I_{ref}$, while the synchronized real video $V_a$ is used only as the evaluation target. We train \newmethod{} on seven GPUs and use one additional GPU for concurrent validation, evaluating checkpoints on the validation set every 2{,}000 training steps. Unless otherwise specified, the 10{,}000-step checkpoint is used for the main comparison and background-preservation experiments, while the training-formulation ablation is evaluated at 4{,}000 steps for all variants.

\subsection{Comparison with Existing Methods}
\label{sec:main-comparison}

We compare \newmethod{} with four publicly available head-swapping baselines: HeadSwap~\cite{lesliezhoa}, the head-swapping branch of REFace~\cite{baliah2024realistic}, Ghost-2.0~\cite{groshev2025ghost2}, and VACE~\cite{vace}. The first three are dedicated identity-swapping methods, while VACE serves as a video-generation baseline. We use the released models with recommended settings and do not retrain them on our paired data. We also compare with HeSer~\cite{shu2022few} and SwapAnyHead~\cite{ji2025controllable} using examples from their papers, as neither provides public code or pretrained models; see Appendix~B of the supplementary material.

\begin{table*}[t]
\centering
\small
\setlength{\tabcolsep}{3.5pt}
\caption{Comparison with publicly available baselines applicable to head swapping on the full-reference benchmark. SR is the success rate of each method; quality metrics, including FID, are computed over each method's usable outputs. }
\label{tab:main_comparison}
\setlength{\tabcolsep}{1.5pt}
\begin{tabular}{lccccccccccccc}
\hline
Method& SR & $\mathit{Face}$ $\uparrow$& $\mathit{Head}$ $\uparrow$& $\mathit{Exp}$ $\uparrow$& $\mathit{Pose(deg^2)}$ $\downarrow$& $\mathit{Key}$ $\downarrow$& $\mathit{Eye}$ $\downarrow$& $\mathit{Mouth}$ $\downarrow$& LPIPS $\downarrow$& SSIM $\uparrow$& PSNR $\uparrow$& FID $\downarrow$& tLPIPS $\downarrow$ \\
\hline
HeadSwap~\cite{lesliezhoa}
& 98.9 & 0.3773 & 0.7850 & 0.7292 & 2.68 & 0.0561 & 0.0822 & 0.0623 & 0.2433 & 0.7293 & 20.59 & 38.97 & 0.0443 \\
REFace (Head)~\cite{baliah2024realistic}
& 98.6 & 0.4526 & 0.5564 & 0.7819 & 1.82 & 0.0395 & 0.0940 & 0.0477 & 0.2195 & 0.7621 & 18.78 & 32.62 & 0.0218 \\
Ghost-2.0~\cite{groshev2025ghost2}
& 98.7 & 0.5708 & 0.7965 & 0.8706 & 0.95 & 0.0345 & 0.0537 & 0.0356 & 0.1981 & 0.7810 & 21.54 & 21.53 & 0.0123 \\
VACE~\cite{vace}
& 100.0 & 0.6639 & 0.8444 & 0.6969 & 8.06 & 0.1747 & 0.1159 & 0.0821 & 0.2142 & 0.7456 & 19.38 & 18.30 & 0.0070 \\
\hline
\textbf{\newmethod{}}
& \textbf{100.0} & \textbf{0.7722} & \textbf{0.8974} & \textbf{0.9063} & \textbf{0.84} & \textbf{0.0255} & \textbf{0.0490} & \textbf{0.0321} & \textbf{0.1205} & \textbf{0.8133} & \textbf{23.98} & \textbf{9.28} & \textbf{0.0055} \\
\hline
\end{tabular}
\end{table*}

\begin{figure*}[t]
  \centering
  \includegraphics[width=1.0\linewidth]{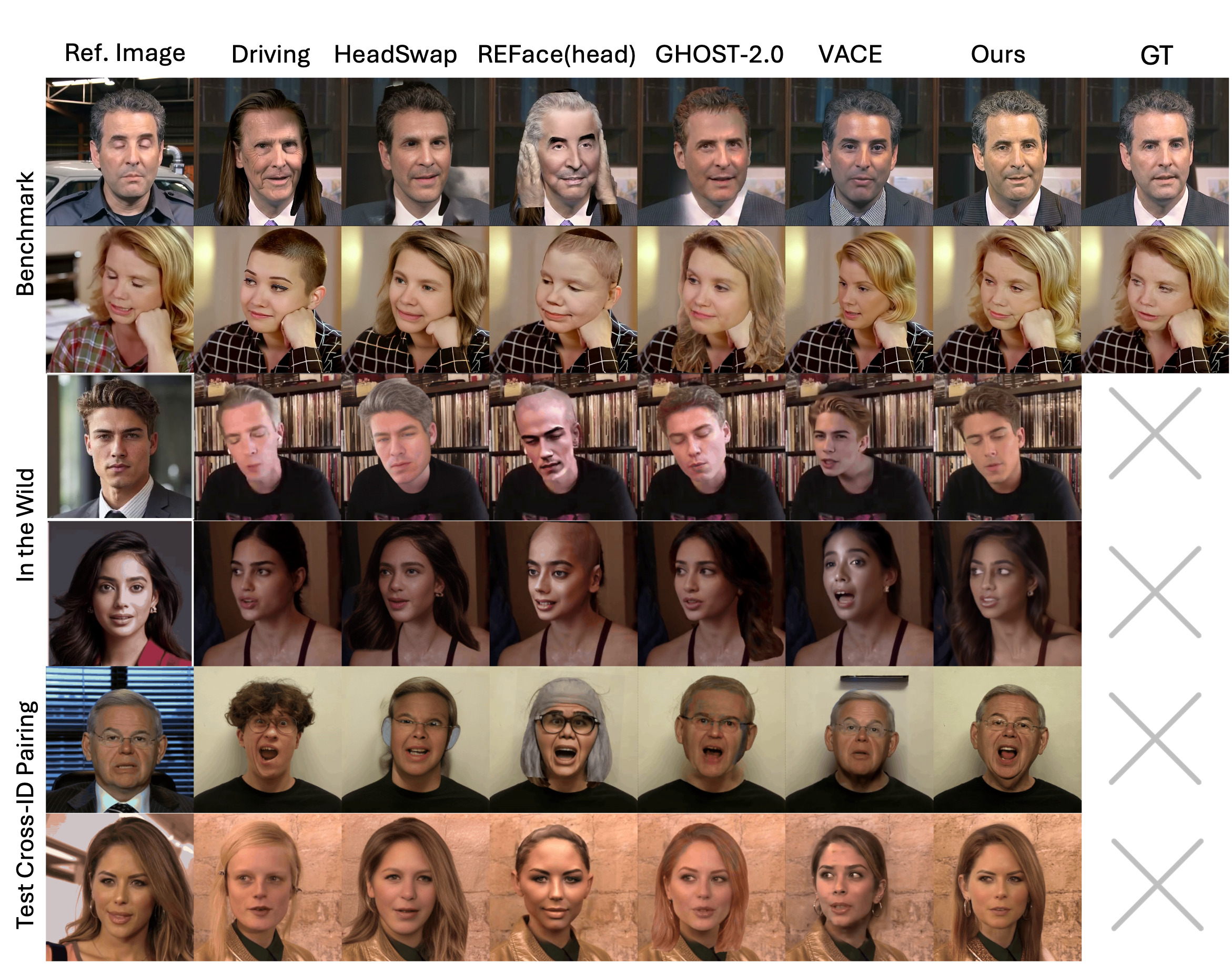}
  \caption{\textbf{Comparison with publicly available methods.} The first two rows show benchmark examples with frame-aligned ground truth; the next two use in-the-wild cases, and the final two are random cross-identity pairings of real benchmark videos. Ground truth is unavailable for the latter four rows. }
  \label{fig:compare}
\end{figure*}

The comparison in \cref{tab:main_comparison} assesses whether the proposed paired data can support a competitive head-swapping system, rather than isolating the effects of training data or architecture. Most released baselines use mask-based same-identity reconstruction and cannot directly consume our cross-identity supervision without modifying their training formulation; we therefore evaluate them as released. DirectSwap performs strongly across identity, motion, and full-reference reconstruction metrics, showing that the paired data can support effective whole-head swapping with a generic video-generation backbone. Relative to the released VACE model, DirectSwap uses the same backbone family at a much smaller scale (1.3B vs.~14B), with higher identity similarity (0.7722 vs.~0.6639), higher expression similarity (0.9063 vs.~0.6969), and lower pose error (0.84 vs.~8.06).

\begin{figure*}[t]
  \centering
  \includegraphics[width=1.0\linewidth]{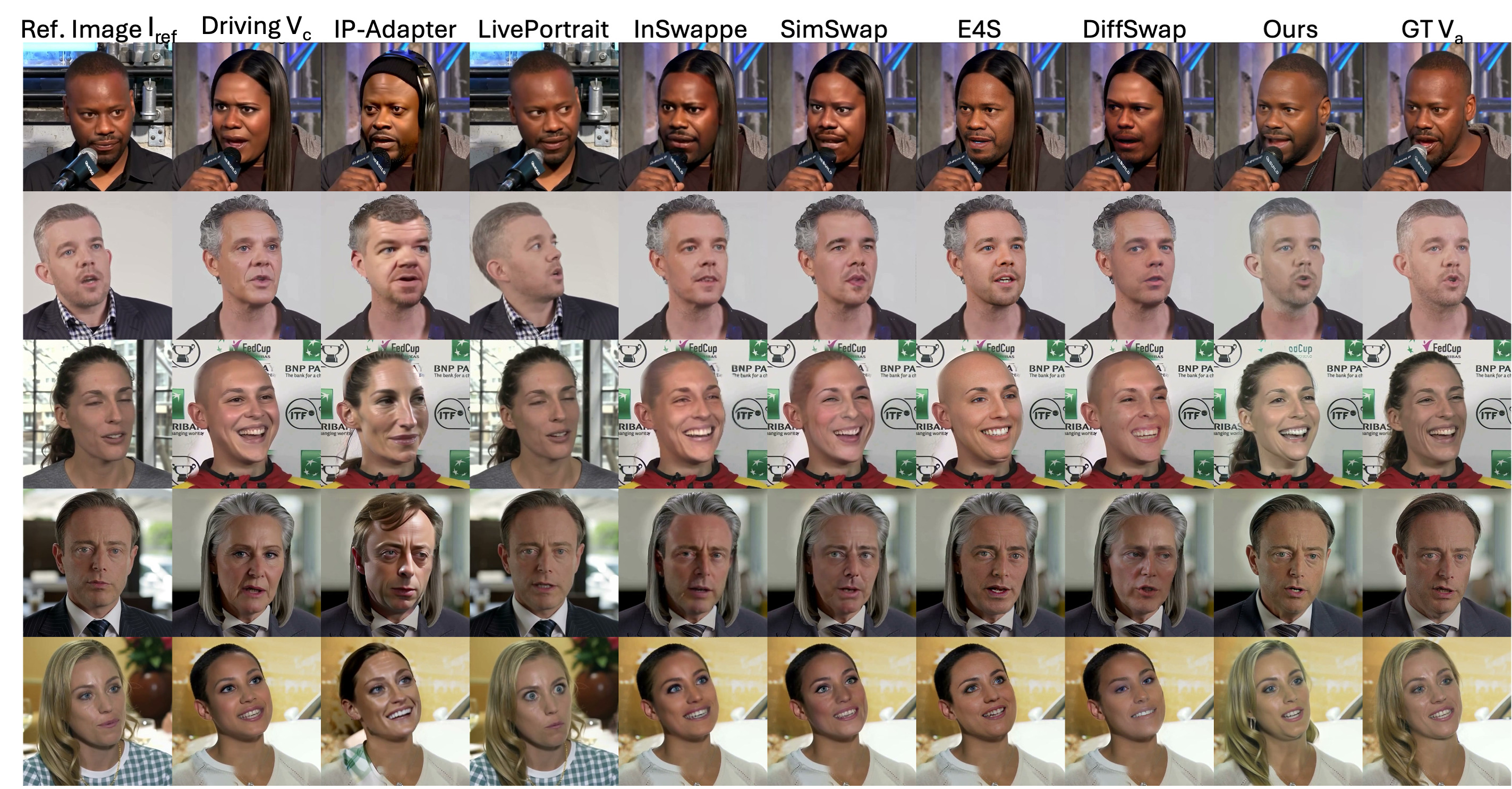}
  \caption{\textbf{Comparison with adjacent task formulations.} We compare IP-Adapter, LivePortrait, and face-swapping methods (InSwapper, SimSwap, E4S, and DiffSwap) on \benchmark{}. Face swapping retains the driving hairstyle and head geometry, whereas \newmethod{} transfers the whole head.}
  \label{fig:faceswap}
\end{figure*}

\noindent\textbf{Adjacent Task Formulations.}
\cref{fig:faceswap} further contrasts \newmethod{} with three adjacent families: subject-driven generation (IP-Adapter~\cite{ye2023ip-adapter}), portrait reenactment (LivePortrait~\cite{guo2024liveportrait}), and face swapping (InSwapper~\cite{haofanwang}, SimSwap~\cite{chen2020simswap}, E4S~\cite{liu2023fine}, DiffSwap~\cite{zhao2023diffswap}). Subject-driven generation reproduces only loose target appearance, reenactment animates the reference portrait without preserving the driving scene, and face swapping changes the inner face while retaining the driving hairstyle and head geometry, which visually separates face swapping from whole-head swapping. Quantitative results for these methods under the same full-reference protocol are reported in Appendix~B of the supplementary material.

\subsection{User Study}
\label{sec:user-study}

We conduct a user study with two goals: verifying that the benchmark comparison in \cref{tab:main_comparison} agrees with human judgment, and testing whether the benchmark trends extend to footage for which no frame-aligned ground truth exists.

\noindent\textbf{Protocol.}
The study comprises 60 cases from three subsets. The \emph{benchmark} subset (30 cases) follows the setting of \cref{tab:main_comparison}. The \emph{in-the-wild} subset (15 cases, 3 references $\times$ 5 driving videos) uses collected out-of-domain footage. The \emph{test cross-ID pairing} subset (15 cases, 3 references $\times$ 5 driving videos) pairs test-split references with test-split driving videos of different identities from $V_a$. Only the benchmark subset has a synchronized real target; the latter two therefore test whether the benchmark trends extend beyond the full-reference setting. For each case, raters view the reference image, driving video, and outputs of all five methods in randomized order with method identities hidden. They mark every output satisfying each of four criteria (identity similarity, expression consistency, blending naturalness, and temporal stability) and additionally select the best overall output, with ``none'' allowed. The study includes 102 raters and 887 case-level judgments. Informed consent was obtained from all participants prior to their participation in the user study. Full protocol and aggregation details are provided in Appendix~E of the supplementary material.

\begin{table*}[t]
\centering
\caption{User study results. ID/Expr/Blend/Temp are acceptance rates (\%); Best is the overall preference rate.}
\label{tab:user_study}
\setlength{\tabcolsep}{2.8pt}
\begin{tabular}{l|ccccc|ccccc|ccccc|ccccc}
\hline
Method & \multicolumn{5}{c|}{Benchmark (30)} & \multicolumn{5}{c|}{In-the-wild (15)} & \multicolumn{5}{c|}{Test cross-ID pairing (15)} & \multicolumn{5}{c}{All (60)} \\
 & ID & Expr & Blend & Temp & Best & ID & Expr & Blend & Temp & Best & ID & Expr & Blend & Temp & Best & ID & Expr & Blend & Temp & Best \\
\hline
HeadSwap~\cite{lesliezhoa} & 23.7 & 39.8 & 5.4 & 1.4 & 0.0 & 53.5 & 36.8 & 3.0 & 0.0 & 0.0 & 10.7 & 11.3 & 0.0 & 0.0 & 0.0 & 27.9 & 31.9 & 3.4 & 0.7 & 0.0 \\
REFace (Head)~\cite{baliah2024realistic} & 1.6 & 17.1 & 0.6 & 0.3 & 0.0 & 1.3 & 12.1 & 0.0 & 0.6 & 0.0 & 0.0 & 11.3 & 0.0 & 0.0 & 0.0 & 1.1 & 14.4 & 0.3 & 0.3 & 0.0 \\
Ghost-2.0~\cite{groshev2025ghost2} & 60.4 & 89.2 & 37.6 & 21.7 & 0.0 & 70.4 & 94.1 & 57.5 & 40.0 & 11.9 & 51.4 & 79.3 & 11.3 & 8.0 & 3.3 & 60.7 & 87.9 & 36.0 & 22.9 & 3.8 \\
VACE~\cite{vace} & 73.4 & 19.9 & 75.8 & 80.7 & 0.7 & 33.3 & 15.3 & 56.6 & 62.1 & 5.3 & 46.5 & 6.7 & 76.7 & 80.1 & 1.3 & 56.6 & 15.4 & 71.3 & 75.9 & 2.0 \\
\textbf{\newmethod{}} & \textbf{99.6} & \textbf{99.6} & \textbf{99.2} & \textbf{99.9} & \textbf{99.3} & \textbf{96.0} & \textbf{96.7} & \textbf{95.5} & \textbf{97.3} & \textbf{82.2} & \textbf{86.7} & \textbf{97.3} & \textbf{98.0} & \textbf{98.0} & \textbf{86.7} & \textbf{95.5} & \textbf{98.3} & \textbf{98.0} & \textbf{98.8} & \textbf{91.9} \\
\hline
\end{tabular}
\end{table*}

\noindent\textbf{Results.}
\cref{tab:user_study} shows that \newmethod{} achieves the highest acceptance rate on every dimension of every subset and is selected as the best output in 91.9\% of all judgments, compared with 3.8\% for the strongest baseline. The human ratings also closely follow the benchmark metrics. VACE performs well in blending and temporal stability but poorly in expression, whereas Ghost-2.0 shows the opposite profile, consistent with their quantitative results in \cref{tab:main_comparison}. Identity judgments show a similarly consistent trend across the baselines.

\newmethod{} remains the most preferred method on both subsets without frame-aligned ground truth, reaching 82.2\% on in-the-wild footage and 86.7\% on test cross-ID pairings. These results show consistent preference trends on cross-identity and out-of-domain footage without frame-aligned ground truth.

\subsection{Masked Reconstruction vs. Mask-Free Paired Training}
\label{sec:mask-ablation}

We compare conventional masked reconstruction with our mask-free cross-identity training formulation. All variants use the same Wan2.1-VACE-1.3B backbone, optimization settings, training split, and number of training steps, but differ in how the training input is constructed. As shown in \cref{fig:datavalid}, the masked variants start from the target video $V_a$, remove the head region, and reconstruct the original $V_a$, using either a segmentation-derived head mask or a coarser rectangular mask. In contrast, \newmethod{} takes the complete cross-identity driving video $V_c$ as input and sets the generation mask to all ones, with the same $V_a$ as the frame-aligned target.

\begin{table}[t]
\centering
\small
\caption{Effect of mask-free paired training.}
\label{tab:mask_ablation}
\setlength{\tabcolsep}{1.5pt}
\begin{tabular}{lcccccc}
\hline
Training
& $\mathit{Face}$ $\uparrow$
& $\mathit{Head}$ $\uparrow$
& $\mathit{Exp}$ $\uparrow$
& $\mathit{Key}$ $\downarrow$
& LPIPS $\downarrow$
& PSNR $\uparrow$ \\
\hline
Head-mask
& 0.6703 & 0.8447 & 0.7400 & 0.0784 & 0.1690 & 21.46 \\
Rect-mask
& 0.6750 & 0.8592 & 0.7264 & 0.1465 & 0.1886 & 20.69 \\
 \textbf{Mask-free (ours)}
& \textbf{0.7676} & \textbf{0.8940} & \textbf{0.8911} & \textbf{0.0272} & \textbf{0.1567} & \textbf{22.12} \\
\hline
\end{tabular}
\end{table}

As shown in \cref{tab:mask_ablation}, mask-free training outperforms both masked variants on every metric. Expression similarity increases from 0.7400/0.7264 to 0.8911, keypoint error decreases by $2.9\times$/$5.4\times$, and identity similarity improves from 0.6703/0.6750 to 0.7676. Under otherwise identical training settings, these results show that consuming the paired data through full-frame cross-identity training is more effective than reducing the task to same-identity masked reconstruction.

\begin{table*}[t]
\centering
\small
\setlength{\tabcolsep}{3.5pt}
\caption{Evaluation on silhouette-change subsets of the test split. The three training-formulation variants use the 4{,}000-step checkpoints from \cref{tab:mask_ablation}, while \newmethod{} uses the final checkpoint from \cref{tab:main_comparison}.}
\label{tab:silhouette}
\begin{tabular}{lcccc|cccc|cccc}
\hline
& \multicolumn{4}{c|}{Short $\rightarrow$ long (76)} & \multicolumn{4}{c|}{Long $\rightarrow$ short (89)} & \multicolumn{4}{c}{No change (875)} \\
Method & $\mathit{Head}$ $\uparrow$ & $\mathit{Exp}$ $\uparrow$ & LPIPS $\downarrow$ & PSNR $\uparrow$
       & $\mathit{Head}$ $\uparrow$ & $\mathit{Exp}$ $\uparrow$ & LPIPS $\downarrow$ & PSNR $\uparrow$
       & $\mathit{Head}$ $\uparrow$ & $\mathit{Exp}$ $\uparrow$ & LPIPS $\downarrow$ & PSNR $\uparrow$ \\
\hline
Head-mask & 0.8133 & 0.7047 & 0.2909 & 15.76 & 0.7760 & 0.6488 & 0.2533 & 18.44 & 0.8484 & 0.7386 & 0.1605 & 21.91 \\
Rect-mask & 0.8427 & 0.6483 & 0.3045 & 15.60 & 0.8123 & 0.6541 & 0.2860 & 17.85 & 0.8591 & 0.7252 & 0.1851 & 20.94 \\
Mask-free (ours) & 0.8719 & 0.8750 & 0.2248 & 18.62 & 0.8834 & 0.8646 & 0.1984 & 21.08 & 0.8965 & 0.8903 & 0.1533 & 22.28 \\
\hline
HeadSwap~\cite{lesliezhoa} & 0.7900 & 0.7154 & 0.3057 & 17.10 & 0.7384 & 0.6861 & 0.2618 & 19.30 & 0.7847 & 0.7285 & 0.2397 & 20.73 \\
Ghost-2.0~\cite{groshev2025ghost2} & 0.8196 & 0.8754 & 0.3031 & 17.16 & 0.7910 & 0.8664 & 0.2249 & 20.46 & 0.7967 & 0.8696 & 0.1942 & 21.74 \\
VACE~\cite{vace} & 0.8089 & 0.6349 & 0.3292 & 15.20 & 0.8325 & 0.6526 & 0.2844 & 17.32 & 0.8459 & 0.6980 & 0.2120 & 19.58 \\
\hline
\textbf{\newmethod{}} & \textbf{0.8929} & \textbf{0.8956} & \textbf{0.1781} & \textbf{20.46} & \textbf{0.8854} & \textbf{0.8821} & \textbf{0.1565} & \textbf{22.71} & \textbf{0.8984} & \textbf{0.9063} & \textbf{0.1179} & \textbf{24.10} \\
\hline
\end{tabular}
\end{table*}

\noindent\textbf{Silhouette-Change Subsets.}
\cref{sec:benchmark-analysis} identifies test clips with substantial driving-target silhouette differences: 76 clips (14 subjects) require short-to-long transitions, corresponding to generation beyond the driving silhouette, while 89 clips (13 subjects) require long-to-short transitions, which require recovery of background, neck, or clothing in vacated regions. The remaining 875 clips form the no-change subset. \cref{tab:silhouette} re-evaluates the three training formulations and the strongest published baselines on these subsets using the same per-subject aggregation. Masked training degrades most on silhouette-changing clips: relative to the mask-free variant, the PSNR gap of head-mask training grows from 0.37\,dB on the no-change subset to 2.86\,dB and 2.64\,dB on short-to-long and long-to-short transitions, respectively, with the same trend in LPIPS. These results show that full-frame cross-identity training is particularly beneficial when the target head must expand beyond or contract within the driving-head silhouette.

\subsection{Background-Preservation Ablation}
\label{sec:bg-ablation}

We evaluate the training-free background-preservation mechanism introduced in \cref{sec:bgpreserve}. All variants use the same \newmethod{} checkpoint and compositing pipeline, with only the inference-time restoration mask varied. We compare the raw output, $D$ alone, $A$ alone, and their union $D\cup A$.

\begin{table}[t]
\centering
\small
\caption{Background-preservation ablation. Segmentation-based restoration uses predicted target-head and vacated-region masks as an external reference.}
\label{tab:bg_ablation}
\setlength{\tabcolsep}{2.2pt}
\begin{tabular}{lcccccc}
\hline
Restoration
& $\mathit{Face}$ $\uparrow$
& $\mathit{Head}$ $\uparrow$
& LPIPS $\downarrow$
& SSIM $\uparrow$
& PSNR $\uparrow$ 
& FID $\downarrow$ \\
\hline
None
& 0.7727 & 0.8993 & 0.1309 & 0.7999 & 23.43 & 12.22 \\
$D$ only
& \underline{0.7307} & 0.8888 & 0.1199 & 0.8159 & 24.07 & 9.12 \\
$A$ only
& 0.7720 & 0.8957 & 0.1251 & 0.8126 & 23.86 & 12.95 \\
\textbf{$D\cup A$}
& 0.7722 & 0.8974 & 0.1205 & 0.8133 & 23.98 & 9.28 \\
\hline
\textcolor{gray}{Segmentation}
& \textcolor{gray}{0.7722} & \textcolor{gray}{0.8975} & \textcolor{gray}{0.1142} & \textcolor{gray}{0.8206} & \textcolor{gray}{24.27} & \textcolor{gray}{8.48}\\
\hline
\end{tabular}
\end{table}

\begin{figure*}[t]
  \centering
  \includegraphics[width=\linewidth]{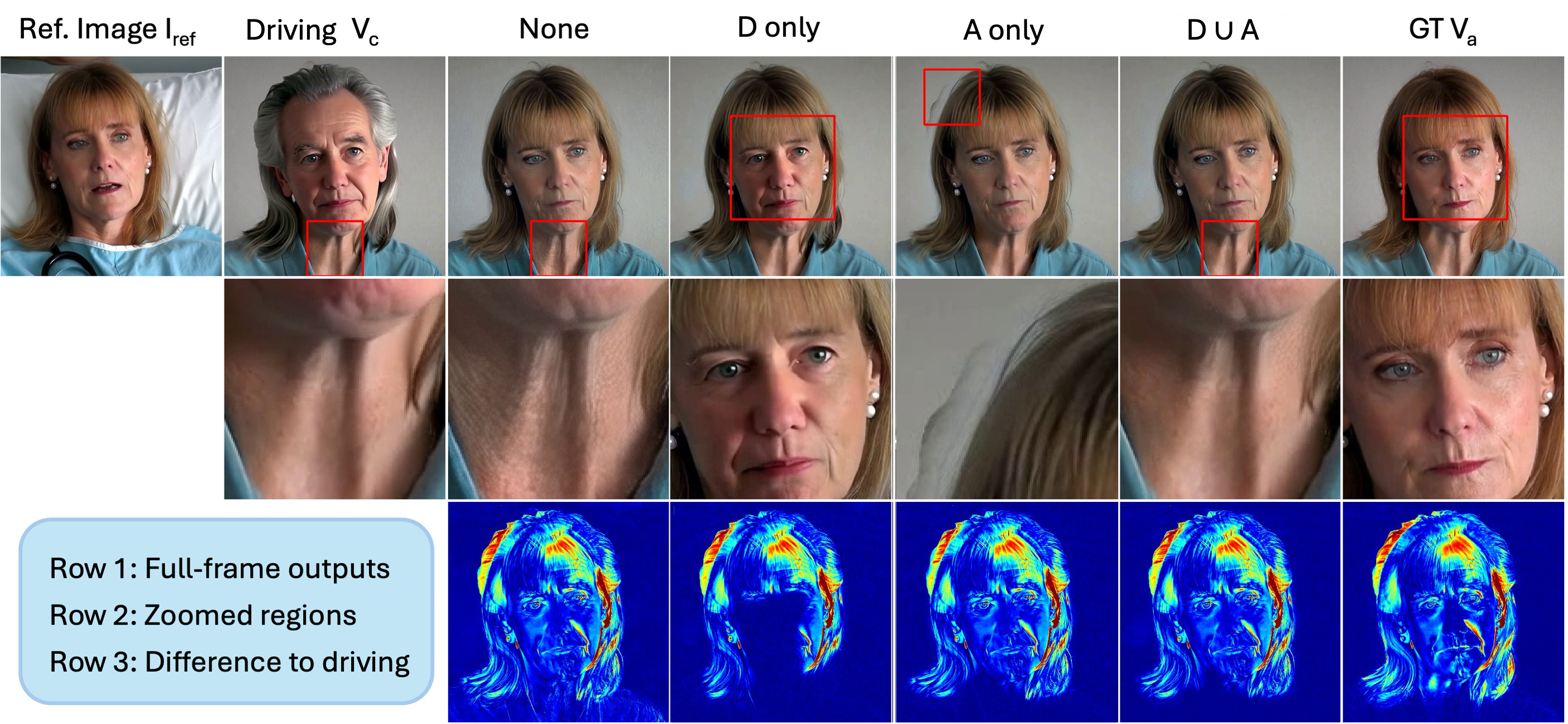}
  \caption{\textbf{Qualitative ablation of the preservation module.}
Rows show full frames, zoomed regions, and absolute differences from the driving video.}
  \label{fig:bg_ablation_vis}
\end{figure*}
As shown in \cref{tab:bg_ablation,fig:bg_ablation_vis}, the two internal signals are complementary. The raw output can alter regions around the neck that should otherwise remain unchanged, illustrating the need for non-head content preservation. Using $D$ alone captures changed regions, including those vacated by the driving head, but may classify low-divergence facial areas as unchanged and restore facial content from the driving video. In contrast, $A$ preserves the reference-conditioned target head but may miss vacated driving-head regions, leaving residual content such as hair. Their union balances the two, retaining both the generated target head and vacated regions while restoring the remaining unchanged non-head content.

\noindent\textbf{Comparison with Segmentation-Based Restoration.}
We additionally compare with a segmentation-based alternative that constructs the edit support from predicted head masks of the generated and driving videos. It achieves slightly better reconstruction and distributional fidelity (LPIPS 0.1142 vs.\ 0.1205; FID 8.48 vs.\ 9.28), while identity and whole-head similarity are essentially identical. In contrast, our internal read-out estimates the restoration support directly from signals already available in the generation process and requires no external parsing model. As video generation moves toward fewer-step and lower-latency inference, avoiding fixed external parsing overhead may become increasingly beneficial.

%% file: sec/7_conclusion.tex
\section{Conclusion}
\label{sec:conclusion}

Video head swapping has long lacked cross-identity paired supervision and frame-aligned ground truth because real footage cannot capture different identities performing the same motion. We address this limitation with an identity-expression decoupled synthesis pipeline that constructs synchronized cross-identity video pairs. These pairs enable \benchmark{}, a cross-identity full-reference benchmark with frame-aligned real targets, and provide direct supervision for training head-swapping models. As one instantiation, \newmethod{} learns from complete cross-identity driving videos; controlled experiments show that this full-frame formulation is more effective than masked same-identity reconstruction, while driving-output divergence and reference attention support training-free preservation of unchanged non-head content at inference. Extensive experiments validate the quality and utility of the paired data, and a 102-rater user study shows that benchmark trends are consistent with human judgment and extend to footage without frame-aligned ground truth.

%% file: sec/supply.tex

\section{Dataset}
\label{sec:dataset}

\subsection{Corpus Composition and Per-Corpus Analysis}
\label{app:corpus}

\subsubsection{Corpus Composition}
\benchmark{} is constructed from HDTF~\cite{zhang2021flow}, VFHQ~\cite{xie2022vfhq}, and FEED~\cite{drobyshev2024emoportraits}. The training split contains 20{,}278 clips from 1{,}345 subjects (11{,}931 HDTF, 4{,}001 VFHQ, and 4{,}346 FEED), while the test split contains 1{,}040 clips from 104 subjects (434 HDTF, 353 VFHQ, and 253 FEED). The 40-clip validation split contains 21 HDTF and 19 VFHQ clips from 16 additional subjects and includes no FEED footage.

\subsubsection{Per-Corpus Expression and Pose Coverage}
\cref{tab:coverage_corpus} shows distinct motion and expression profiles across the three source corpora. HDTF is dominated by talking-head speech: 95.6\% of its test clips contain speech activity, with particularly high mouth-opening and squint frequencies. FEED shows the opposite profile: only 2.8\% of its clips contain speech, but it has the highest rates of extreme expressions (73.2\%) and large head rotations (34.4\%), together with strong brow-raise, smile, and surprise coverage. VFHQ occupies an intermediate regime, with moderate speech activity, expression intensity, and large-pose frequency.

The corpora also differ in the tails of their head-pose distributions. HDTF has a narrow yaw range but pronounced downward pitch; VFHQ shows broader roll variation, and FEED reaches the largest bidirectional yaw excursions. Together, these complementary regimes broaden the motion and expression coverage of the combined benchmark.

\begin{table*}[t]
\centering
\small
\caption{Per-corpus expression, clip statistics, and head-pose statistics on the \benchmark{} test split. Head-pose rows report frame-level percentiles.}
\label{tab:coverage_corpus}
\begin{tabular}{lccc}
\hline
& HDTF (434) & VFHQ (353) & FEED (253) \\
\hline
Mouth open            & 44.0\% & 17.3\% & 22.4\% \\
Surprise              & 26.7\% & 10.2\% & 31.2\% \\
Frown                 & 44.2\% & 29.7\% & 18.0\% \\
Smile                 & 30.6\% & 51.6\% & 54.0\% \\
Brow raise            & 31.1\% & 33.1\% & 55.2\% \\
Pucker                & 48.2\% & 38.2\% & 38.8\% \\
Squint                & 76.0\% & 70.8\% & 36.4\% \\
Eyes closed           & 26.5\% & 22.7\% & 8.4\% \\
\hline
Neutral-only clips    & 3.5\%  & 4.0\%  & 13.2\% \\
Speech activity       & 95.6\% & 58.4\% & 2.8\% \\
Extreme expr.         & 39.2\% & 50.4\% & 73.2\% \\
Large pose            & 9.4\%  & 14.7\% & 34.4\% \\
\hline
Yaw $[p_5/p_{50}/p_{95}]$
& $-9.7^\circ/+1.0^\circ/+7.5^\circ$
& $-30.2^\circ/-2.1^\circ/+20.1^\circ$
& $-30.4^\circ/+1.5^\circ/+30.4^\circ$ \\
Pitch $[p_5/p_{50}/p_{95}]$
& $-22.4^\circ/-1.3^\circ/+4.6^\circ$
& $-9.0^\circ/+3.3^\circ/+15.9^\circ$
& $-16.9^\circ/-7.1^\circ/+4.9^\circ$ \\
Roll $[p_5/p_{50}/p_{95}]$
& $-4.1^\circ/+2.7^\circ/+9.2^\circ$
& $-13.1^\circ/+0.5^\circ/+14.1^\circ$
& $-14.4^\circ/-0.9^\circ/+10.0^\circ$ \\
\hline
Blink events / clip   & 2.7  & 1.7  & 0.3 \\
\hline
\end{tabular}
\end{table*}

\subsubsection{Expression Category Definitions}
For the category-level statistics in Table~III of the main paper and \cref{tab:coverage_corpus}, we derive eight interpretable expression categories from the MediaPipe blendshape coefficients. A clip is assigned to a category when its category-specific activation criterion is satisfied in at least two frames, except for eye closure, which requires at least three consecutive closed-eye frames. A clip is labeled \emph{neutral} when none of the eight categories is triggered over its entire duration; the neutral rate is thus the complement of the category assignments, not a ninth criterion. The complete category definitions are listed in \cref{tab:expression_definition}.

Two additional measures capture temporal dynamics. A clip is labeled as containing \emph{speech activity} if the smoothed \texttt{jawOpen} signal shows at least three open-close cycles with peak prominence $>0.06$ and standard deviation $>0.03$, thereby distinguishing repeated jaw motion from sustained mouth opening. A \emph{blink event} is counted whenever $\operatorname{mean}(\texttt{eyeBlink}^{\mathrm{L,R}})$ crosses $0.5$ upward, complementing the eyes-closed category by measuring discrete rather than sustained closures.

\begin{table*}[t]
\centering
\small
\caption{Blendshape-based definitions of the eight expression categories used
for coverage analysis.}
\label{tab:expression_definition}
\setlength{\tabcolsep}{3.5pt}
\begin{tabular}{lll}
\hline
Category & Blendshape signal & Activation criterion \\
\hline
Mouth open  & \texttt{jawOpen} & $>0.40$ \\
Surprise    & \texttt{browInnerUp} & $>0.40$, and (mean(\texttt{eyeWide}$^{\mathrm{L,R}}$)$>0.15$ or \texttt{jawOpen}$>0.35$) \\
Frown       & mean(\texttt{browDown}$^{\mathrm{L,R}}$) & $>0.35$ \\
Smile       & mean(\texttt{mouthSmile}$^{\mathrm{L,R}}$) & $>0.50$ \\
Brow raise  & mean(\texttt{browOuterUp}$^{\mathrm{L,R}}$) & $>0.50$ \\
Pucker      & \texttt{mouthPucker} & $>0.60$ \\
Squint      & mean(\texttt{eyeSquint}$^{\mathrm{L,R}}$) & $>0.55$ \\
Eyes closed & mean(\texttt{eyeBlink}$^{\mathrm{L,R}}$) & $>0.60$, $\geq 3$ consec.\ frames \\
\hline
\end{tabular}
\end{table*}

\begin{figure*}[t]
\centering
\includegraphics[width=\linewidth]{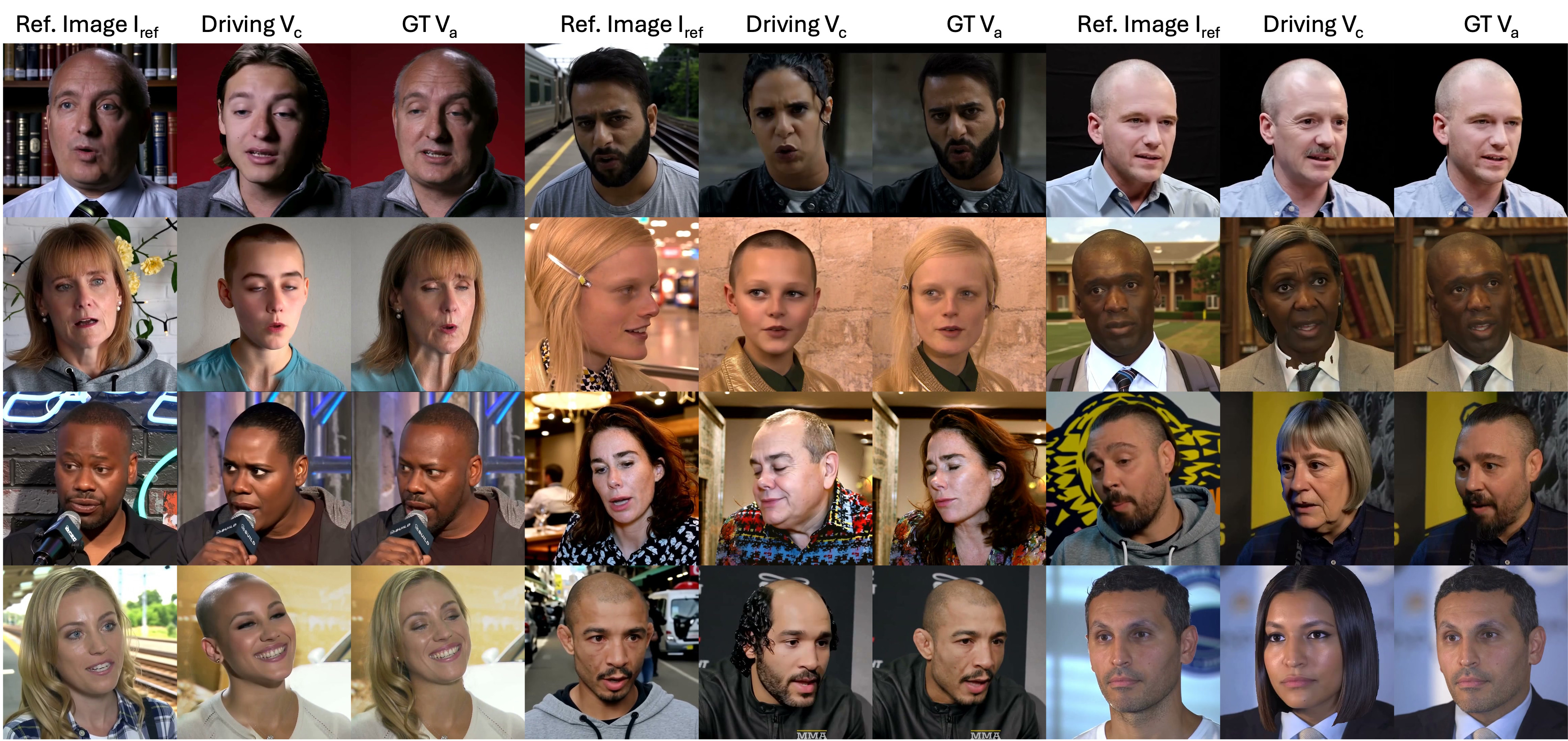}
\caption{\textbf{Representative samples from \benchmark{}.} Each triplet shows the driving video $V_c$, target-identity reference $I_{ref}$, and synchronized real target $V_a$. The examples illustrate the diversity of identity, expression, head pose, hairstyle, and head-silhouette changes in the test split.}
\label{fig:benchmark_gallery}
\end{figure*}

\noindent\textbf{Benchmark Visualization.}
\cref{fig:benchmark_gallery} shows representative samples from the test split, covering diverse identities, expressions, head poses, hairstyles, and silhouette changes. Each example contains the driving video $V_c$, target reference $I_{ref}$, and synchronized real target $V_a$.

\subsection{Synthesis Implementation Details}
\label{app:synth-impl}
All video-inpainting passes of the synthesis pipeline, including identity inpainting, background refinement, and the non-head inpainting of the reference-image branch (Section~III-A of the main paper), run as ComfyUI workflows on Wan2.1-T2V-14B~\cite{wan2025}, accelerated with the LightX2V CFG-step-distillation LoRA (rank 32, bf16), which distills classifier-free guidance into the model and enables few-step sampling. With this acceleration, one inpainting pass over a 121-frame clip takes approximately 2.5 minutes on a single GPU.

\subsection{Comparison with Keypoint-Conditioned Synthesis}
\label{app:synth-keypoint}

We also evaluate a keypoint-conditioned alternative that synthesizes a new identity directly from the facial-landmark sequence of the real video $V_a$, without transplanting pixels from $V_a$. Using the same source corpora and backbone family, we generate 8{,}066 candidates from HDTF and VFHQ with landmark-conditioned Wan2.1-VACE at $512{\times}512$ and approximately 121 frames per clip. Applying the same acceptance criteria and thresholds as in Section~III-A of the main paper, only \textbf{0.4\%} of these candidates pass the filter, compared with \textbf{78\%} for our synthesis.

\section{Experiments}
\label{sec:more_experiments}
\subsection{Training Details}
\label{app:training}

DirectSwap is fine-tuned from Wan2.1-VACE-1.3B with the full DiT trainable and the Wan-VAE and text encoder frozen. Training uses the 20{,}278 paired clips at $512\times512$ resolution with mixed clip-length buckets of 17, 49, 81, and 121 frames and per-GPU batch sizes of 4, 2, 1, and 1, respectively. The reference image $I_{ref}$ is prepended as a reference latent frame following VACE, and the generation mask is set to all ones.

We use 8-bit AdamW with learning rate $10^{-5}$, $\beta_1{=}0.9$, $\beta_2{=}0.95$, weight decay $0.01$, and gradient clipping at $1.0$. Training runs on seven NVIDIA A100 GPUs, with one additional GPU used for asynchronous validation. Checkpoints are evaluated every 2{,}000 steps on the 40-clip validation split. Unless otherwise stated, the 10,000-step checkpoint is used for the reported results; the controlled training-formulation ablation uses the 4,000-step checkpoint for all variants, as specified in the main paper.

Inference uses 30-step flow matching with sampling shift 16 and classifier-free guidance scale 5.0. A 121-frame clip takes approximately 3 minutes on a single 80\,GB GPU. The background-preservation read-out in Section~V-B adds no additional model forward pass beyond the final RGB compositing.

\subsection{Per-Corpus FID}
\label{app:fid-corpus}

\cref{tab:fid_corpus} breaks the overall FID of Table~IV of the main paper down by source corpus. Following the protocol of Table~IV in the main paper, FID for each method is computed between all frames of its usable outputs and the corresponding real target frames. \newmethod{} attains the lowest FID in every corpus. The breakdown also localizes baseline failures that the overall number hides: REFace degrades sharply on the studio-captured FEED corpus (105.47 vs.\ 26.91 on HDTF), whereas VACE and DirectSwap show smaller increases in FID on the same footage.

\begin{table}[t]
\centering
\small
\caption{FID by source corpus. Each method is evaluated over its usable clips.}
\label{tab:fid_corpus}
\setlength{\tabcolsep}{4.5pt}
\begin{tabular}{lcccc}
\hline
Method & Overall$\downarrow$ & HDTF$\downarrow$ & VFHQ$\downarrow$ & FEED$\downarrow$ \\
\hline
HeadSwap~\cite{lesliezhoa} & 38.97 & 42.48 & 51.47 & 58.89 \\
REFace (Head)~\cite{baliah2024realistic} & 32.62 & 26.91 & 32.91 & 105.47 \\
Ghost-2.0~\cite{groshev2025ghost2} & 21.53 & 14.87 & 28.35 & 57.43 \\
VACE~\cite{vace} & 18.30 & 14.84 & 26.70 & 45.73 \\
\hline
\textbf{\newmethod{}} & \textbf{9.28} & \textbf{7.36} & \textbf{14.51} & \textbf{22.17} \\
\hline
\end{tabular}
\end{table}

\subsection{Results by Head-Pose Bucket}
\label{app:pose-bucket}

\cref{tab:pose_bucket} splits the test set by the large-head-rotation criterion of Section~IV-C of the main paper (at least one frame with $|\mathrm{yaw}|>30^\circ$ or $|\mathrm{pitch}|>25^\circ$): 179 clips (17.2\%, 20 identities) qualify as large-pose, and the remaining 861 clips (95 identities) as routine; identities with clips in both buckets contribute to both, and per-identity aggregation is applied within each bucket. Facial identity similarity declines for every method at large pose, but DirectSwap maintains strong performance under large pose: its whole-head similarity is essentially unchanged (0.900 vs.\ 0.899), its large-pose facial identity (0.680) still exceeds every baseline's \emph{routine}-pose score, and its perceptual advantage widens (LPIPS 0.133 vs.\ 0.210 for the best baseline).

\begin{table*}[t]
\centering
\small
\setlength{\tabcolsep}{4pt}
\caption{Results by head-pose bucket (per-identity aggregation within each bucket). Large pose: $\geq$1 frame with $|\mathrm{yaw}|>30^\circ$ or $|\mathrm{pitch}|>25^\circ$ (179 clips); Routine: the remaining 861 clips.}
\label{tab:pose_bucket}
\begin{tabular}{l ccccc | ccccc}
\hline
& \multicolumn{5}{c|}{Large pose (179)} & \multicolumn{5}{c}{Routine (861)} \\
Method & $\mathit{Face}$ $\uparrow$ & $\mathit{Head}$ $\uparrow$ & $\mathit{Exp}$ $\uparrow$ & LPIPS $\downarrow$ & PSNR $\uparrow$ & $\mathit{Face}$ $\uparrow$ & $\mathit{Head}$ $\uparrow$ & $\mathit{Exp}$ $\uparrow$ & LPIPS $\downarrow$ & PSNR $\uparrow$ \\
\hline
HeadSwap~\cite{lesliezhoa} & 0.3412 & 0.7952 & 0.6927 & 0.2598 & 19.85 & 0.3825 & 0.7842 & 0.7283 & 0.2405 & 20.70 \\
REFace (Head)~\cite{baliah2024realistic} & 0.3964 & 0.5775 & 0.7658 & 0.2238 & 18.21 & 0.4579 & 0.5529 & 0.7811 & 0.2199 & 18.78 \\
Ghost-2.0~\cite{groshev2025ghost2} & 0.5557 & 0.8029 & 0.8583 & 0.2115 & 21.13 & 0.5731 & 0.7972 & 0.8703 & 0.1965 & 21.63 \\
VACE~\cite{vace} & 0.5746 & 0.8580 & 0.6641 & 0.2101 & 19.13 & 0.6752 & 0.8458 & 0.6979 & 0.2143 & 19.45 \\
\hline
\textbf{\newmethod{}} & \textbf{0.6802} & \textbf{0.8997} & \textbf{0.8882} & \textbf{0.1333} & \textbf{23.37} & \textbf{0.7830} & \textbf{0.8988} & \textbf{0.9072} & \textbf{0.1191} & \textbf{24.09} \\
\hline
\end{tabular}
\end{table*}

\subsection{Comparison with Face Swapping, Reenactment, and Subject-Driven Generation}
\label{app:faceswap}

The main comparison (Table~IV of the main paper) is restricted to methods that attempt whole-head swapping. For completeness, \cref{tab:faceswap_comparison} evaluates representative methods from three adjacent families under the identical protocol: subject-driven generation (IP-Adapter~\cite{ye2023ip-adapter}), portrait reenactment (LivePortrait~\cite{guo2024liveportrait}, driving the reference image with the driving video's motion), and face swapping (InSwapper~\cite{haofanwang}, SimSwap~\cite{chen2020simswap}, E4S~\cite{liu2023fine}, DiffSwap~\cite{zhao2023diffswap}). Benchmark visualizations are shown in Fig.~8 of the main paper.

The results illustrate the characteristic trade-offs of these adjacent task formulations under our full-reference protocol. IP-Adapter performs generic subject-conditioned generation without frame-aligned supervision, reproducing only loose target appearance (identity similarity 0.2091). LivePortrait instead animates the reference portrait itself, which naturally favors identity similarity to the reference but does not preserve the driving scene and framing. Face-swapping methods primarily modify the facial region while retaining the driving hairstyle and surrounding head geometry: InSwapper reaches a facial identity similarity (0.7817) comparable to ours, and this family attains the best temporal stability because only a small facial region is modified. However, whole-head similarity remains substantially lower (0.5250-0.6882 vs.\ 0.8974), reflecting the retained driving hairstyle and head geometry and thus the distinction between face and head swapping. DirectSwap provides a strong balance between whole-head identity transfer and preservation of the driving content.

\begin{table*}[t]
\centering
\small
\setlength{\tabcolsep}{3.5pt}
\caption{Comparison with subject-driven, reenactment, and representative face-swapping methods on the full-reference benchmark.}
\label{tab:faceswap_comparison}
\begin{tabular}{lcccccccccccc}
\hline
Method& SR & $\mathit{Face}$ $\uparrow$& $\mathit{Head}$ $\uparrow$& $\mathit{Exp}$ $\uparrow$& $\mathit{Pose(deg^2)}$ $\downarrow$& $\mathit{Key}$ $\downarrow$& $\mathit{Eye}$ $\downarrow$& $\mathit{Mouth}$ $\downarrow$& LPIPS $\downarrow$& SSIM $\uparrow$& PSNR $\uparrow$& tLPIPS $\downarrow$ \\
\hline
IP-Adapter~\cite{ye2023ip-adapter}
& 100.0 & 0.2091 & 0.6715 & 0.7119 & 7.67 & 0.0797 & 0.0916 & 0.0660 & 0.2207 & 0.7423 & 18.97 & 0.0183 \\
LivePortrait~\cite{guo2024liveportrait}
& 100.0 & \textbf{0.8226} & 0.8873 & 0.7865 & 1.90 & 0.2457 & 0.0886 & 0.0632 & 0.5637 & 0.4368 & 11.98 & 0.0114 \\
InSwapper~\cite{haofanwang}
& 99.8 & 0.7817 & 0.6882 & 0.8549 & 0.90 & 0.0299 & 0.0554 & 0.0359 & 0.2039 & 0.7905 & 20.40 & 0.0047 \\
SimSwap~\cite{chen2020simswap}
& 99.5 & 0.5163 & 0.5866 & 0.8552 & 1.02 & 0.0309 & 0.0613 & 0.0356 & 0.1982 & 0.7809 & 20.01 & \textbf{0.0043} \\
E4S~\cite{liu2023fine}
& 93.5 & 0.4278 & 0.5575 & 0.8250 & 1.19 & 0.0367 & 0.0638 & 0.0438 & 0.1975 & 0.7653 & 19.78 & 0.0077 \\
DiffSwap~\cite{zhao2023diffswap}
& 94.1 & 0.4342 & 0.5250 & 0.8044 & 1.79 & 0.0320 & 0.0690 & 0.0454 & 0.1969 & 0.7757 & 19.76 & 0.0258 \\
\hline
\textbf{\newmethod{}}
& \textbf{100.0} & 0.7722 & \textbf{0.8974} & \textbf{0.9063} & \textbf{0.84} & \textbf{0.0255} & \textbf{0.0490} & \textbf{0.0321} & \textbf{0.1205} & \textbf{0.8133} & \textbf{23.98} & 0.0055 \\
\hline
\end{tabular}
\end{table*}

\subsection{Qualitative Comparison}
\label{app:qualcomp}

\begin{figure*}[t]
  \centering
  \includegraphics[width=1.0\linewidth]{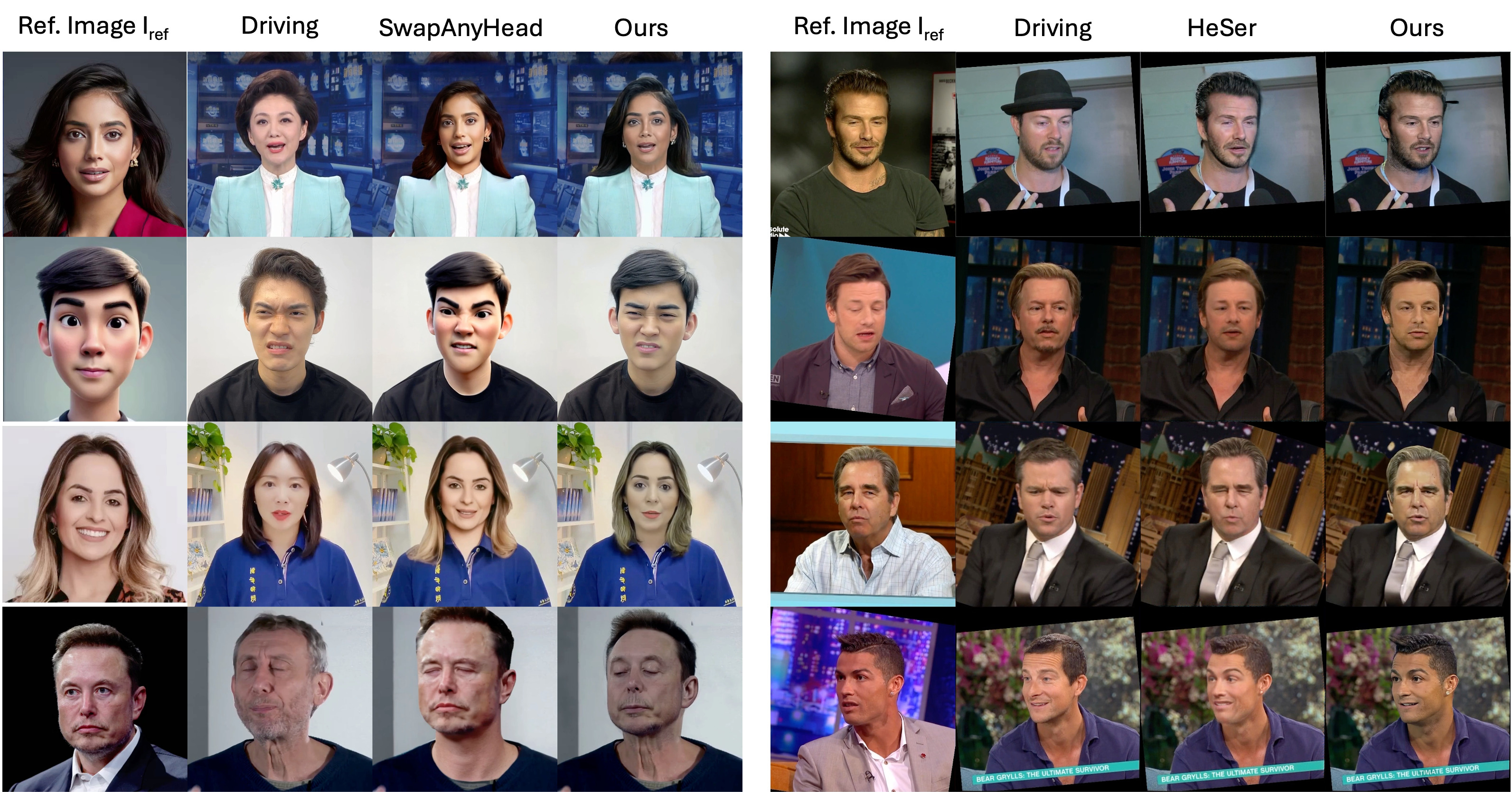}
  \caption{Comparison with SwapAnyHead~\cite{ji2025controllable} and HeSer~\cite{shu2022few}. }
  \label{fig:swapanyhead}
\end{figure*}

\noindent\textbf{Comparison with HeSer and SwapAnyHead.}
Neither HeSer~\cite{shu2022few} nor SwapAnyHead~\cite{ji2025controllable} releases code or trained models, so a benchmark evaluation is not possible. We instead run our model on the qualitative examples published in their papers and compare our results with the reported results (\cref{fig:swapanyhead}). In these examples, DirectSwap shows better preservation of facial expression and closer agreement with the input video's color tone and illumination.

\section{Background-Preservation Read-Out: Design and Parameter Selection}
\label{app:params}

All validation-dependent design choices and thresholds in the background-preservation read-out (Section~V-B of the main paper) are selected exclusively on the 40-clip validation split defined in Section~IV-A of the main paper; the 1{,}040-clip test split is never used for parameter selection. Other implementation constants are fixed as specified below. All calibration results are measured on the \emph{primary} checkpoint, i.e., the DirectSwap model used in the experiments. Where robustness across models is examined, we additionally report an \emph{independent} checkpoint, a separately trained DirectSwap model with the same training recipe but a different training mask, and the \emph{base} model, the pretrained Wan2.1-VACE-1.3B without fine-tuning.

For calibration and evaluation only, the reference edit support is constructed as the union of the head masks of the driving and target frames, obtained using the same segmentation pipeline as in dataset construction. These masks are never used during inference. We report the estimated edit support against this reference using \emph{IoU}, \emph{recall} (the fraction of the reference edit support covered by the estimated mask), \emph{precision} (the fraction of the estimated mask lying inside the reference edit support), and their harmonic mean \emph{F1}. All quantities are computed per clip and averaged over clips. Although IoU and F1 are the standard overall measures, the two error types are not symmetric for compositing: a precision loss merely leaves a small background region to the generator, whereas a recall loss reintroduces driving-head content. We therefore prefer a slightly lower IoU or F1 when it buys a clear gain in recall.

\subsection{Which Layers Carry the Localization}
\label{app:params-layers}

Measured per layer over the validation clips at the final denoising step, localization quality is sharply band-structured (\cref{tab:params_layers}). Three characteristic regimes emerge. \emph{Diffuse}: early layers attend to the reference nearly uniformly, providing little spatial discrimination. \emph{Inverted}: several later and VACE-control blocks attend more strongly to the background than to the head (head/background ratio $<1$), so including them can bias the aggregate response toward the background. \emph{Usable}: layers L8-L26 exhibit the clearest and most consistent localization.

\begin{table}[t]
\centering
\small
\caption{Localization by layer band (validation split, final denoising step). IoU is the best achievable against the reference edit region when scanning thresholds; ratio is head-to-background attention mass.}
\label{tab:params_layers}
\begin{tabular}{lcc}
\hline
Band & Best IoU & Head/bg ratio \\
\hline
L0-L7 & 0.06-0.41 & 0.77-1.89 \\
\textbf{L8-L26} & \textbf{0.41-0.76} & \textbf{2.0-10.4} \\
L27-L29 & 0.06-0.17 & 0.52-0.92 \\
VACE branch (15 blocks) & 0.02-0.32 & 0.38-1.42 \\
\hline
\end{tabular}
\end{table}

\subsection{Frame-Wise Normalization}
\label{app:params-frame}

Attention magnitudes are not directly comparable across latent frames. We empirically observe consistently larger responses in the first latent frame than in later frames. Using a single quantile over the entire clip therefore over-selects the first frame and under-selects later frames. We instead compute the top-$30\%$ quantile independently for each latent frame, so that every frame contributes the same fraction of selected tokens. This improves localization consistently; for example, at $\tau_A=0.2$, recall increases from 0.851 to 0.867.

\subsection{Layer Aggregation}
\label{app:params-agg}

Given the per-frame responses above, we combine layers by voting. Each layer independently marks its top $30\%$ of tokens in every latent frame, and the binary maps are averaged over L8-L26. Thus, the resulting value at each token is simply the fraction of selected layers that agree on that location and requires no additional per-clip normalization.

\cref{tab:params_agg} compares several aggregation strategies across three checkpoints. Voting yields the highest head-to-background contrast on all three checkpoints and the best localization on the unfine-tuned base model. Although mean aggregation achieves slightly higher oracle IoU on the two fine-tuned checkpoints, voting produces substantially stronger head-to-background separation. We therefore adopt voting for its more robust threshold separation across model variants.

\begin{table}[t]
\centering
\small
\setlength{\tabcolsep}{3.5pt}
\caption{Aggregation over L8-L26: best IoU / head-to-background ratio on three checkpoints. \emph{Primary} is the DirectSwap checkpoint used in the experiments; \emph{Independent} is a separately trained DirectSwap model with the same training recipe but a different training mask; \emph{Base} is the pretrained Wan2.1-VACE-1.3B without fine-tuning.}
\label{tab:params_agg}
\begin{tabular}{lccc}
\hline
Aggregation & Primary & Independent & Base \\
\hline
Mean & \textbf{0.739} / 5.09 & \textbf{0.701} / 3.91 & 0.423 / 1.64 \\
Median & 0.735 / 5.59 & 0.694 / 4.25 & 0.385 / 1.67 \\
Geometric mean & 0.735 / 5.20 & 0.696 / 3.98 & 0.389 / 1.69 \\
\textbf{Vote (top 30\%)} & 0.714 / \textbf{10.32} & 0.681 / \textbf{6.84} & \textbf{0.458} / \textbf{2.17} \\
All 45 layers, mean & 0.633 / 2.27 & 0.624 / 2.48 & 0.246 / 1.34 \\
\hline
\end{tabular}
\end{table}

\subsection{Divergence Construction}
\label{app:params-div}
The channel-averaged driving-output latent difference is computed at the final diffusion denoising step and smoothed independently per frame with a 2D Gaussian ($\sigma=1$ latent cell), without temporal accumulation. We use $1.2m_t$ as the frame-relative center and $0.4m_t$ as the sigmoid scale, both fixed implementation constants. The former sets the binary decision boundary above typical within-frame variation, while the latter controls only response softness; specifically, $D>0.5$ is equivalent to $\tilde d_t>1.2m_t$.

\subsection{Thresholds}
\label{app:params-thr}

We select the two thresholds sequentially on the validation split. We first calibrate the reference-attention matte using $A$ alone to determine $\tau_A$, then fix $\tau_A=0.2$ and tune $\tau_D$ for the union $A{>}0.2 \lor D{>}\tau_D$.

\begin{figure*}
  \centering
  \includegraphics[width=1.0\linewidth]{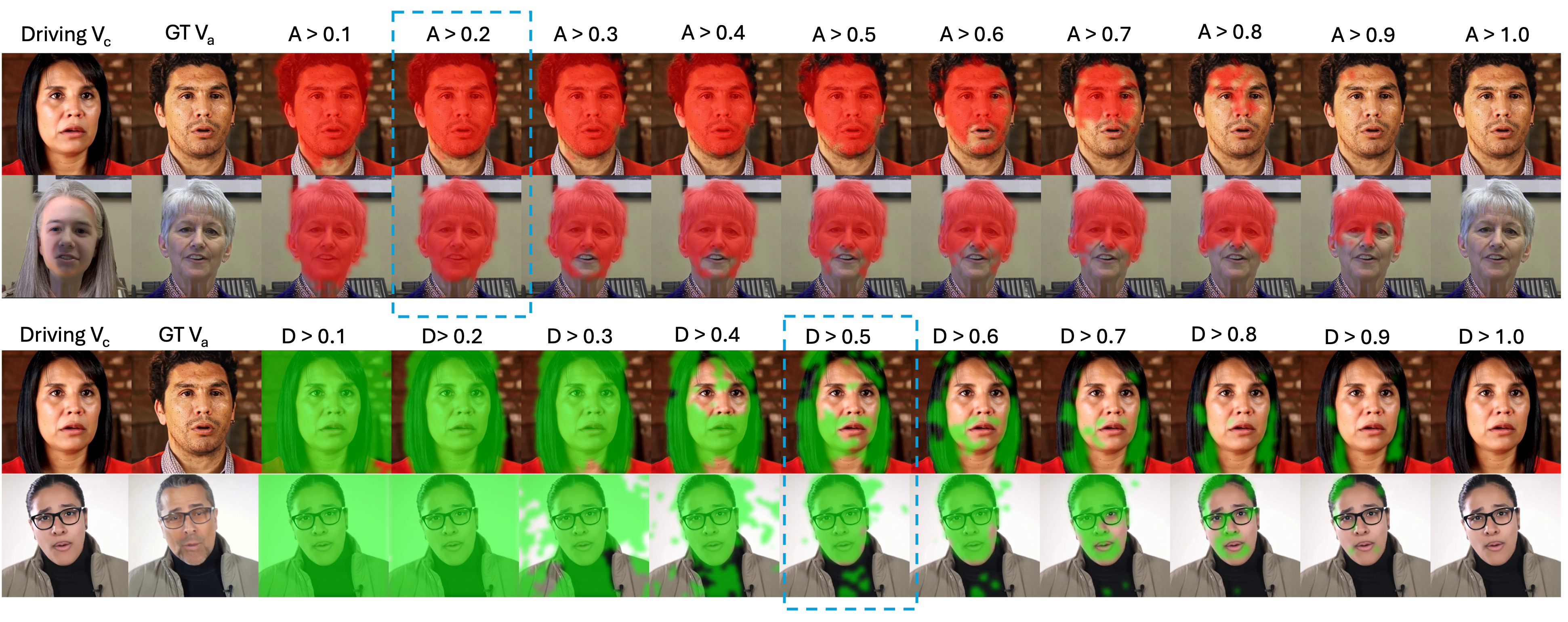}
  \caption{Qualitative visualization of the threshold sweep on four representative validation clips. Dashed boxes mark the selected values ($\tau_A{=}0.2$, $\tau_D{=}0.5$). }
  \label{fig:md_ablation}
\end{figure*}

\noindent\textbf{$\tau_A=0.2$.}
As shown in \cref{tab:params_taum}, F1 is nearly flat between $\tau_A=0.2$ and $0.3$ (0.835 vs.\ 0.839), and IoU likewise (0.727 vs.\ 0.732), so the overall measures do not separate the two settings. The choice is therefore driven by recall, which we prioritize because false negatives inside the edited head reintroduce driving-face content during compositing: $\tau_A=0.2$ retains a recall of 0.867 versus 0.828 at 0.3. Lowering the threshold further to 0.1 yields only a modest recall gain (0.904) while F1 drops to 0.808. We therefore use $\tau_A=0.2$ throughout all experiments.

\begin{table}[t]
\centering
\small
\setlength{\tabcolsep}{5pt}
\caption{Threshold sweeps on the validation split. Top: $\tau_A$ with $A$ alone. Bottom: $\tau_D$ for the union $D{>}\tau_D \lor A{>}0.2$; $\tau_D=1.0$ disables $D$. Selected values in bold.}
\label{tab:params_taum}
\begin{tabular}{lcccc}
\hline
$\tau_A$ & IoU $\uparrow$ & Recall $\uparrow$ & Precision $\uparrow$ & F1 $\uparrow$ \\
\hline
0.1 & 0.693 & 0.904 & 0.769 & 0.808 \\
0.2 & 0.727 & 0.867 & 0.837 & 0.835 \\
0.3 & 0.732 & 0.828 & 0.877 & 0.839 \\
0.4 & 0.715 & 0.780 & 0.906 & 0.825 \\
0.5 & 0.677 & 0.720 & 0.930 & 0.793 \\
0.7 & 0.540 & 0.560 & 0.962 & 0.661 \\
\hline
$\tau_D$ & IoU $\uparrow$ & Recall $\uparrow$ & Precision $\uparrow$ & F1 $\uparrow$ \\
\hline
0.3 & 0.575 & 0.979 & 0.591 & 0.712 \\
0.4 & 0.673 & 0.963 & 0.700 & 0.794 \\
0.5 & 0.723 & 0.946 & 0.763 & 0.832 \\
0.6 & 0.743 & 0.933 & 0.795 & 0.846 \\
0.8 & 0.754 & 0.913 & 0.825 & 0.853 \\
1.0 (no $D$) & 0.727 & 0.867 & 0.837 & 0.835 \\
\hline
\end{tabular}
\end{table}

\noindent\textbf{$\tau_D=0.5$.}
With $A$ fixed at $\tau_A=0.2$, adding $D$ at $\tau_D=0.5$ raises recall from 0.867 to 0.946 at precision 0.763, while IoU (0.723 vs.\ 0.727) and F1 (0.832 vs.\ 0.835) remain close to the $A$-only values. IoU and F1 peak at a stricter $\tau_D=0.8$ (0.754 and 0.853), but that setting recovers fewer vacated-head regions not covered by $A$ (recall 0.913), and residual driving hair in these regions is far more visible than the precision lost at 0.5. Lower thresholds, in turn, sacrifice precision quickly (0.700 at 0.4, F1 0.794). We therefore use $\tau_D=0.5$ throughout.

\subsection{Mask Denoising}
\label{app:params-speck}

After thresholding, we denoise the binary response independently in each frame
using connected components. We retain the largest component, together with components covering at least four latent cells or lying within two latent cells of the largest component. At $512{\times}512$ resolution, these thresholds correspond to $1024\,\mathrm{px}^2$ and $32$\,px, respectively. These values follow from the latent-grid resolution rather than separate tuning. This
denoising leaves recall unchanged (0.867) while raising precision from 0.825 to 0.837 and F1 from 0.827 to 0.835 at $\tau_A=0.2$.

\subsection{Morphological Post-Processing and Feathering}
\label{app:params-feather}

After denoising, we apply per-frame binary closing with a $5{\times}5$ square structuring element, followed by hole filling. The binary edit support is then converted to a soft blending matte using outward Gaussian feathering with $\sigma=8$ pixels: the mask is dilated by $2\sigma$ before Gaussian smoothing and combined with the original support. Values below $0.02$ and above $0.98$ are snapped to 0 and 1, respectively.

\subsection{Read-Out Step}
\label{app:params-step}

We extract the reference-attention response at the final denoising step. Earlier steps provide less discriminative localization, while aggregating responses across multiple steps yields only marginal gains on the validation split. Moreover, the thresholds used by the restoration module are calibrated on the final-step response. We therefore use the final denoising step without temporal accumulation throughout all experiments.

\subsection{Prompt Invariance}
\label{app:params-prompt}

We test whether the reference-attention matte depends on the inference text prompt. The 40 validation clips are regenerated with four prompts: the default inference prompt, two unrelated prompts, and the empty string. For each run, we measure the localization of the reference self-attention against the ground-truth head region.

\begin{table}[t]
\centering
\small
\caption{Prompt-invariance analysis on the validation split. Reference-attention localization remains stable across unrelated and empty prompts.}
\label{tab:params_prompt}
\begin{tabular}{lcc}
\hline
Prompt & Ref.\ attn.\ mean IoU & min IoU \\
\hline
``A person is talking.'' & 0.845 & 0.535 \\
``A dog is running.'' & 0.843 & 0.557 \\
``A landscape with mountains.'' & 0.844 & 0.539 \\
empty prompt & 0.846 & 0.545 \\
\hline
\end{tabular}
\end{table}

As shown in \cref{tab:params_prompt}, the reference-attention localization is nearly unchanged across prompts, with mean IoU varying only from 0.843 to 0.846. The same behavior persists even with an empty prompt, indicating that the matte does not rely on person-related text semantics. We observe the same trend on an independently trained checkpoint. These results indicate that the spatial localization of $A$ is largely insensitive to text semantics and requires no prompt engineering.

\section{Effect of Large Pitch Mismatch on Identity Evaluation}
\label{app:calibration}

We further examine the effect of large pitch mismatch on identity evaluation and assess the benefit of frame-synchronized targets over a fixed reference image. The study uses only real footage from our corpus and involves no generative model. We evaluate AdaFace~\cite{kim2022adaface} and the whole-head similarity~\cite{wang2026head} metric used in the main paper, and additionally include ArcFace~\cite{deng2019arcface} as an independent face-recognition embedding to verify that the observed behavior is not specific to AdaFace.

For each large-pitch query frame, we compare identity similarity against four types of references: (i) a near-frontal frame of the same subject, (ii) a pose-matched frame of the same subject, (iii) a pose-matched frame of a different subject, and (iv) a near-frontal frame of a different subject. The same-subject references are drawn from different clips to avoid trivial frame matching. This design allows us to separate the effects of identity and pose mismatch under both genuine and impostor comparisons.

\begin{figure*}
  \centering
  \includegraphics[width=1.0\linewidth]{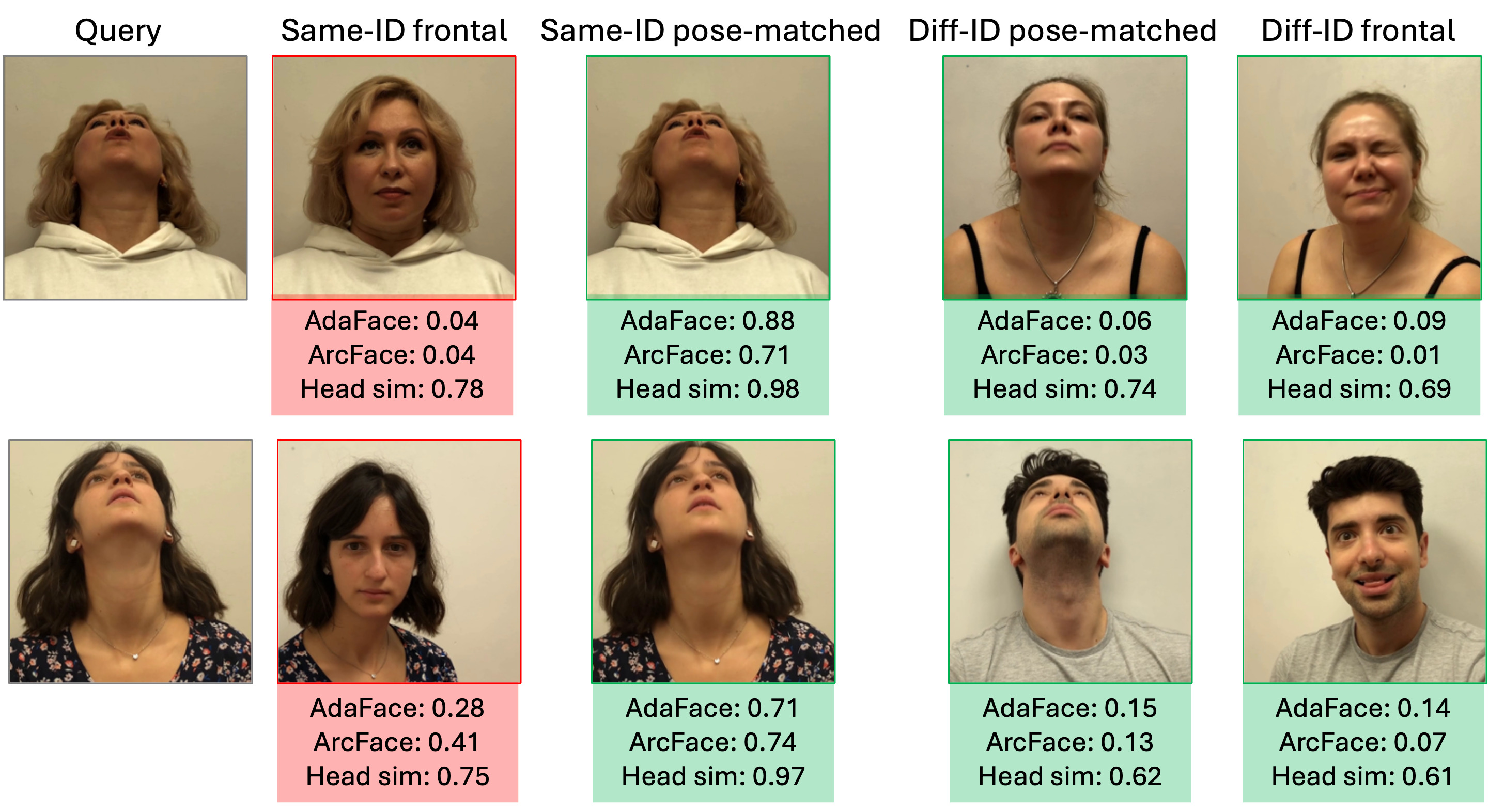}
  \caption{\textbf{Effect of large pitch mismatch on identity evaluation.} For each query, we compare same- and different-identity references under both near-frontal and pose-matched conditions. Under AdaFace, ArcFace, and whole-head similarity, the same-identity score drops sharply when the reference is near frontal, but recovers under pose matching, with pose-matched same-identity references restoring a clear margin over different-identity references. This shows that large pitch mismatch can confound single-reference identity evaluation and motivates frame-synchronized full-reference comparison.}
  \label{fig:calib_examples}
\end{figure*}

As illustrated in \cref{fig:calib_examples}, a large pitch mismatch can substantially reduce same-identity similarity. A near-frontal reference of the same subject may receive a low score, whereas a pose-matched same-identity reference restores substantially higher similarity under AdaFace, ArcFace, and whole-head similarity. In particular, pose matching restores a clear margin between the same-identity reference and different-identity references.

\section{User Study Protocol}
\label{app:userstudy}

\noindent\textbf{Evaluation Cases.}
The study contains 60 cases from three subsets. The \emph{benchmark} subset contains 30 cases sampled from the test split, with coverage of large head poses, silhouette changes, strong expressions, and routine cases. The \emph{in-the-wild} subset contains 15 cases formed by three out-of-domain reference identities and five out-of-domain driving videos. The \emph{test cross-ID pairing} subset contains 15 cases formed by three test-split references and five test-split driving videos of different identities from $V_a$. Only the benchmark subset has a synchronized real target; the latter two therefore evaluate performance beyond the full-reference setting. Each case contains the outputs of all five compared methods.

\noindent\textbf{Interface and Questions.}
Each case displays the reference image, driving video, and five anonymized method outputs simultaneously, with the output order randomized independently for every case. All videos can be replayed freely. For the first four questions, raters mark every output that satisfies the specified criterion, with ``none'' allowed:
\begin{enumerate}
\item ``Same head as the reference photo (incl.\ hairstyle \& head shape)?'' (\emph{identity similarity})
\item ``Follows the input video's expression?'' (\emph{expression consistency})
\item ``Looks like a real, unedited video?'' (\emph{blending naturalness})
\item ``Temporally stable --- no flicker or appearance drift?'' (\emph{temporal stability})
\item ``Best head swap overall? (single choice)''
\end{enumerate}

The first four questions correspond to identity similarity, expression consistency, blending naturalness, and temporal stability, respectively. Question~5 asks for the single best overall output, with ``none'' also allowed.

\noindent\textbf{Forms and Raters.}
The 60 cases are distributed over ten forms. Each form contains five or six unique cases drawn from the three subsets, together with three benchmark cases shared across all ten forms. Each form is completed by at least ten raters, yielding 102 raters and 887 case-level judgments in total. The raters were undergraduate, master's, and Ph.D.\ students. Participation was voluntary, and no personally identifying information was collected or stored.

\noindent\textbf{Rater consistency.}
All ten rater groups selected \newmethod{} as the best output on all three shared cases. Across the method $\times$ dimension cells of these cases, the spread between the highest- and lowest-scoring form is 13.5 points on average (median 10.0), and 79\% of cells vary by at most 20 points, suggesting that the overall acceptance patterns are not dominated by a particular rater group.

\noindent\textbf{Aggregation.}
Because the three shared cases are rated by all ten groups, we first compute each score within a case and then average over cases, so that every case has equal weight and the shared cases do not disproportionately affect the reported rates. 

\section{Release}
\label{sec:release}

To support reproducibility and future research, we plan to release the following resources:

\begin{enumerate}
\item The complete code and scripts for constructing the proposed cross-identity paired dataset.
\item The generated dataset, including \TrainNum{} training pairs, 40 validation
pairs, and \TestNum{} test pairs, available under a research-use agreement.
\item The full implementation of \newmethod{} and the trained checkpoints, with
checkpoint access provided under the same research-use agreement.
\end{enumerate}

\noindent\textbf{Distribution.}
The release follows the terms of the source corpora and underlying footage. HDTF~\cite{zhang2021flow} is distributed by its authors as a CC~BY~4.0 URL list, whereas VFHQ~\cite{xie2022vfhq} and FEED~\cite{drobyshev2024emoportraits} are distributed for non-commercial research. Because our synthesized pairs retain content from the original footage, the processed training and test sets will be provided under a signed research-only agreement with a takedown policy, rather than as an unrestricted public release. The trained checkpoints will be distributed under the same agreement. Any applicable consent withdrawal or valid takedown request will be propagated to our release.

\section{Social Impact}
\label{sec:social}

Head swapping has potential applications in personalized video creation, digital avatars, virtual production, and film post-production. It may also support privacy-oriented content creation by replacing a person's visible head with a synthetic identity, although other cues such as voice, body appearance, and scene context may still reveal personal information.

At the same time, realistic identity manipulation can be misused for impersonation, fraud, or misinformation. We therefore restrict access to the trained checkpoints and processed datasets to researchers under a signed research-only agreement. The release also follows the usage conditions of the source corpora and includes a takedown policy for content depicting real individuals, as described in Appendix~\ref{sec:release}.

We also assess detectability with seven off-the-shelf DeepfakeBench detectors~\cite{DeepfakeBench_YAN_NEURIPS2023} trained on FaceForensics++~\cite{roessler2019faceforensicspp}. Without adaptation, they achieve video-level AUCs of 0.71 for distinguishing raw outputs from real videos and 0.65 after background preservation, indicating nontrivial cross-generator detectability but insufficient reliability as a standalone safeguard. To encourage transparent use, our released code will also support labeling generated outputs as AI-generated.

\section{Limitations}
\label{sec:limit}

\begin{figure}[t]
  \centering
  \includegraphics[width=1.0\linewidth]{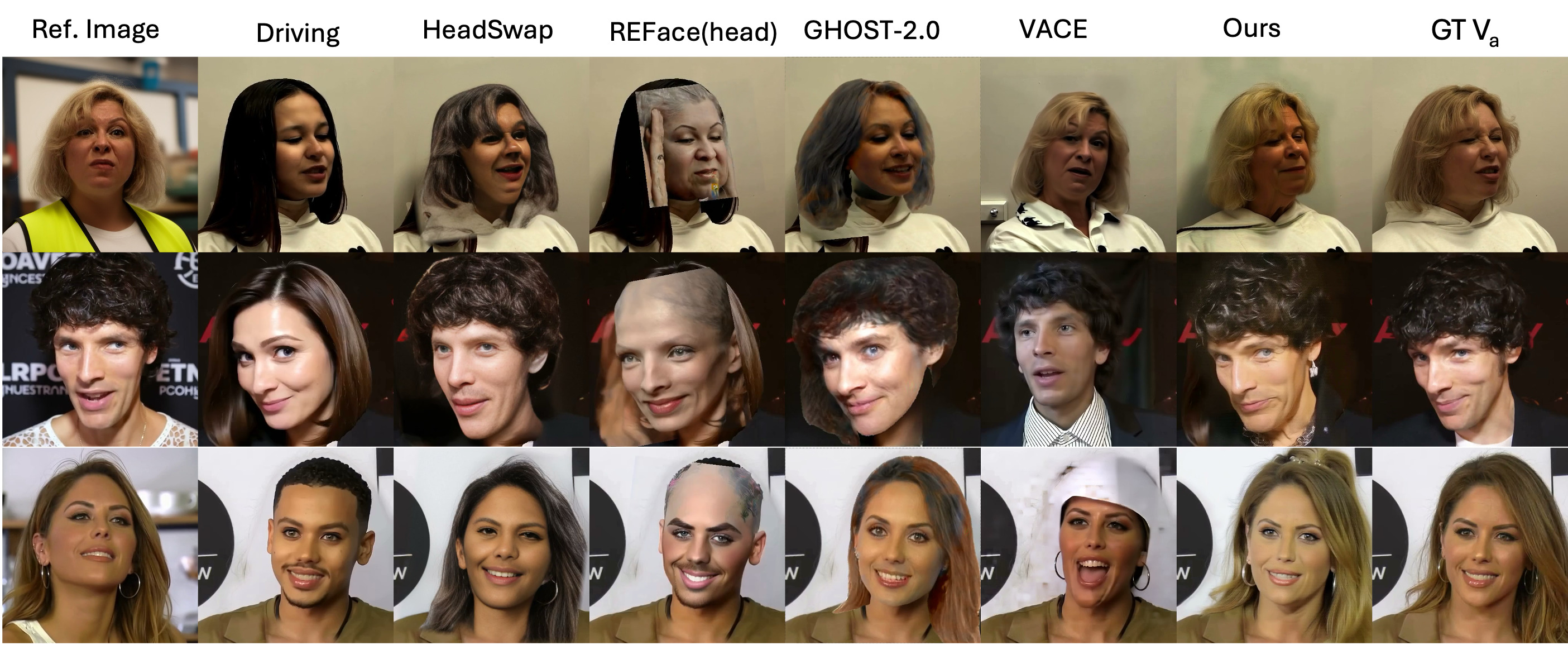}
  \caption{ \textbf{Representative failure cases on \benchmark{}.} \textbf{Top:} the generated identity does not fully match the reference. \textbf{Middle:} residual driving-hair content remains after the swap. \textbf{Bottom:} challenging hair regions contain additional generated textures.
}
  \label{fig:failure}
\end{figure}

Although \newmethod{} performs strongly on the proposed benchmark, several limitations remain. First, whole-head appearance transfer is not yet perfect even for examples within the benchmark distribution. As shown in \cref{fig:failure}, the generated identity may deviate from the reference, residual driving-hair content can remain around the target hairstyle, and additional textures may appear in challenging hair regions. These cases suggest that the current paired supervision and model still leave room for improving fine-grained identity and hairstyle fidelity.

A second limitation arises from the construction of the paired training data. To preserve synchronized facial dynamics, our synthesis pipeline retains localized expression-bearing regions, including the eyebrows, eyes, nose, and mouth, from the real video. Although these regions undergo geometric perturbation and subsequent processing through the video-generation pipeline, some synthesized pairs may still retain noticeable local appearance similarity.

This local overlap does not imply that \newmethod{} simply copies the driving facial components at inference. The model is trained to condition on an independently specified target reference, and our cross-identity and in-the-wild results show that the generated facial appearance can adapt to a different reference identity while preserving the driving expression and motion. Nevertheless, reducing such local overlap in the paired data remains desirable.

A promising direction is to bootstrap pair construction with a trained head-swapping model. Since the current model can transfer independently chosen reference identities onto driving videos, its outputs could be combined with our existing synchronization and quality filters to construct new paired data without explicitly retaining facial components from the original video. This may further reduce local appearance overlap and improve the diversity of paired supervision.

\section{Use of Vision-Language and Language Models}
\label{sec:llm_usage}

A vision-language model (Qwen2.5-VL-72B-Instruct~\cite{qwen2.5-VL}) is used in the data-generation pipeline to extract scene attributes, including background and lighting information, for conditioning identity synthesis. In addition, a large language model is used during manuscript preparation for language polishing and to assist with the implementation of some evaluation scripts. All AI-assisted code was reviewed, verified, and executed by the authors, and all experimental settings, analyses, results, and scientific conclusions were determined and validated by the authors.